\pdfoutput=1
\documentclass{article}

\usepackage[eandd, final]{neurips_2026}

\usepackage[utf8]{inputenc}
\usepackage[T1]{fontenc}
\usepackage{amsmath}
\usepackage{amsfonts}
\usepackage{booktabs}
\usepackage{graphicx}
\usepackage{microtype}
\usepackage{nicefrac}
\usepackage{xcolor}
\usepackage{hyperref}
\usepackage{fontawesome5}
\usepackage{url}
\usepackage{tikz}
\usetikzlibrary{arrows.meta,positioning,fit,backgrounds,calc}

\newcommand{\mbench}{M4Bench}

\newcommand{\mimic}{MIMIC-IV}
\newcommand{\eicu}{eICU}
\newcommand{\gptmini}{GPT~5.4~Mini}
\newcommand{\gptfull}{GPT~5.5}
\newcommand{\noskill}{\textsc{no-skill}}
\newcommand{\withskill}{\textsc{with-skill}}
\newif\ifresulthighlight
\resulthighlightfalse
\definecolor{resultyellow}{rgb}{1,1,0.55}
\newcommand{\resultnum}[1]{%
  \ifresulthighlight
    \colorbox{resultyellow}{\strut #1}%
  \else
    #1%
  \fi}

\newcommand{\releasetag}{m4bench-neurips2026-camera-ready}
\newcommand{\releaseurl}{https://github.com/hannesill/m4/tree/m4bench-neurips2026-camera-ready/benchmark}

\title{\mbench: Evaluating Procedural Specification for Clinical EHR Derivation Agents}

\author{%
  Hannes Ill$^{1,*}$, Rafi Al Attrach$^{1}$, Rajna Fani$^{1}$, Ahram Han$^{1,2}$, \\
  \textbf{Ryohei Yamamoto$^{1}$, Marlene Louisa Moerig$^{1,3}$, Dukyong Yoon$^{1,4}$,
  Leo Anthony Celi$^{1,5}$} \\[1.5ex]
  $^{1}$Laboratory for Computational Physiology, Massachusetts Institute of Technology, USA \\
  $^{2}$Department of Surgery, Seoul National University Hospital, Korea \\
  $^{3}$Charit\'e, University Medicine Berlin, Germany \\
  $^{4}$Department of Biomedical Systems Informatics, Yonsei University College of Medicine, Korea \\
  $^{5}$Division of Pulmonary, Critical Care and Sleep Medicine, \\
  Beth Israel Deaconess Medical Center, USA \\[1.5ex]
  $^{*}$Corresponding author: \texttt{illh@mit.edu} \\[0.3ex]
  \href{\releaseurl}{\faGithub\ \ \texttt{github.com/hannesill/m4}}
}

\begin{document}

\maketitle

\begin{abstract}
Clinical research over electronic health records depends on derived variables whose definitions sit outside the schema, in item codes, time windows, missingness rules, and output-grain choices that change cohorts and downstream conclusions.
We introduce \mbench{}, 28 clinical derivation tasks across 15 task families on \mimic{} and \eicu{}, to test whether AI coding agents can reproduce such derivations and how clinician-reviewed agent skills change their results.
Its controls decompose the value of skills into specification completeness, delivery format, and schema familiarity.
In a Docker-isolated Codex campaign (\gptmini{} and \gptfull{}), skills raise reward, the mean match rate over scored output columns, by \resultnum{0.136} over the no-skill baseline, with \resultnum{26} of 28 tasks improving.
Specification completeness accounts for most of this gain: on an eight-task control subset, an operational spec alone raises reward by \resultnum{0.162}, and skills add only \resultnum{0.024} on top, a difference indistinguishable from zero.
On a ten-task subset, decoy skills score like no skill and raw reference SQL nearly matches targeted skills, arguing against generic context or the skill format as the explanation.
Schema obfuscation and restructuring preserve \resultnum{86\%} and \resultnum{70\%} of the skill gain (\gptmini{}, 26 \mimic{} tasks), a descriptive robustness check rather than a bound on memorization.
A descriptive Claude Code replication on all 28 tasks (Opus~4.7 and Sonnet~4.6, 560 runs) is directionally consistent (\resultnum{+0.120}, \resultnum{21} of 28 tasks improving).
These gains are partial agreement with the reference, not completed tasks: only \resultnum{25} of 280 with-skill and \resultnum{13} of 280 no-skill Codex runs pass every strict check, and skills barely change which rows agents return (key F1 \resultnum{+0.003}).
Skills are released as a versioned, clinician-reviewable markdown delivery format, with a family-level reference-validity audit.
\mbench{} releases tasks, skills, evaluation code, and sanitized per-run evidence as an open-source benchmark~\href{\releaseurl}{\faGithub}.

\end{abstract}

\section{Introduction}

\begin{figure*}[t]
  \centering
  \includegraphics[width=\textwidth]{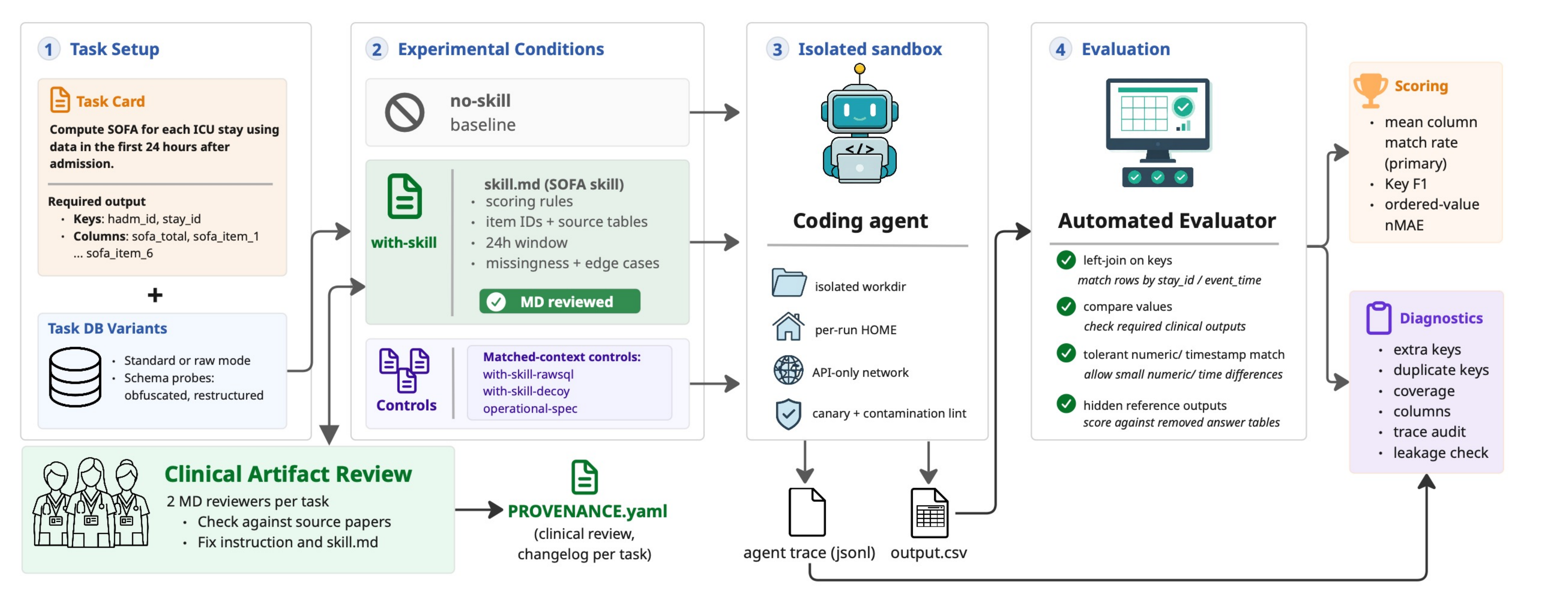}
  \caption{\mbench{} evaluation pipeline. Each task pairs a task card with a DuckDB clone whose reference targets are removed. The primary \noskill{}--\withskill{} comparison and matched controls test specification completeness, reference-code structure, and schema familiarity under sandboxed execution and contamination auditing.}
  \label{fig:pipeline}
\end{figure*}

Observational clinical research using electronic health records (EHRs) commonly relies on derived variables: severity scores such as Sequential Organ Failure Assessment (SOFA)~\citep{vincent1996sofa} and Acute Physiology Score III (APS-III)~\citep{knaus1991apacheiii}, organ-dysfunction measures such as Kidney Disease: Improving Global Outcomes (KDIGO) acute kidney injury staging~\citep{kdigo2012aki} and Model for End-Stage Liver Disease (MELD)~\citep{kamath2001meld}, and cohort-defining phenotypes such as suspected infection and Sepsis-3~\citep{singer2016sepsis3}. These variables are rarely available as simple columns. They are reconstructed from raw tables whose clinical meaning is distributed across item identifiers, units, documentation practices, time windows, and dataset-specific conventions, and the implementation choices behind a single derivation change cohort membership, score distributions, and downstream conclusions in observational studies~\citep{johnson2017reproducibility,johnson2018mimiccode,ostropolets2023reproducible}.

Modern agents can inspect schemas, execute SQL, debug code, and iterate on tests, and recent work measures their ability to translate clinical questions into SQL or to interact with EHR data~\citep{lee2022ehrsql,shi2024ehragent,jiang2025medagentbench}. Yet clinical derivation tasks often require procedural knowledge that is invisible from schema structure alone. A SOFA implementation, for example, needs the score definition plus how vasopressors are encoded, how oxygenation variables are paired, how ventilation status changes respiratory scoring, and how missing components are handled in a specific database release. This procedural knowledge is held by clinicians and embedded in community concept code such as the MIT-LCP MIMIC-Code repository~\citep{johnson2018mimiccode}, but no benchmark we are aware of represents it as an auditable agent artifact.

\mbench{} studies this gap as an evaluation problem. Rather than asking whether an agent can answer clinical questions in free text, \mbench{} asks whether a coding agent can reproduce declared EHR research derivations and whether explicit procedural specifications change that behavior. The benchmark is motivated by a broader goal: enabling clinicians and clinical researchers to convert tacit procedural knowledge about EHR derivations into versioned, auditable artifacts that agents can use during implementation. The central question is how human domain expertise should be represented and evaluated for supervised research workflows.

The primary specification format is an agent \emph{skill}: a markdown procedural artifact, in the spirit of Agent Skills~\citep{anthropic2025agentskills}, that summarizes the clinical rules, item identifiers, time windows, and output grain for one task family. Skills are reusable procedural artifacts and contain no hidden reference outputs. Each task and associated artifact is reviewed by two clinicians. This review supports face validity, output-schema consistency, and clinical-definition provenance.

The benchmark is designed around a controlled ablation (Figure~\ref{fig:pipeline}). In the \noskill{} condition, an agent receives the task instruction and database access, but no curated derivation procedure. This condition represents schema-plus-task access, not a fully specified clinical protocol baseline. In the \withskill{} condition, the same agent also receives the task-specific procedural skill. The submitted CSV is evaluated column by column against hidden reference outputs. The estimand is the change in agreement with declared reference implementations when curated clinical implementation knowledge is supplied.

\paragraph{Claims and non-claims.}
\mbench{} estimates the native-schema \withskill{}--\noskill{} reward difference for the same task, database, model, and execution protocol, with tasks weighted equally after model-level averaging.
The tested procedural context comes from public MIT-LCP concept code, eICU conventions, clinician review, and task adaptations.
Scores measure agreement with declared public-code reference derivations, not clinical correctness: clinician review flagged, for example, that broad culture inclusion can overcall suspected infection, that SOFA defaults missing organ components to zero, and that the vasopressor mapping omits angiotensin~II (Appendix Table~\ref{tab:clinical_reference_audit}). Clinician-adjudicated and multiple accepted targets are the priority for future versions.
The evidence is \mimic{}-dominant: 26 of 28 tasks use \mimic{}, and the two \eicu{} tasks are cross-database probes, not evidence of database or hospital generality.
\mbench{} does not identify a causal effect of clinician review, provider-general behavior, memorization-free reasoning, or a unique effect of the skill format.

This paper makes three contributions:

\begin{enumerate}
    \item A benchmark object: 28 clinical EHR derivation tasks across 15 task families over \mimic{} and \eicu{}, with clinician-reviewed skills, OS-level isolation, contamination-lint, and sanitized per-run audit logs plus a family-level reference-validity audit so that benchmark targets and known caveats are inspectable.
    \item A decomposition apparatus: decoy-skill, raw-SQL, operational-spec, and schema-perturbation controls separating specification completeness from delivery format, reference-code structure, and schema familiarity.
    \item A Codex finding under that apparatus, with a directionally consistent but descriptive full-coverage Claude replication: on an eight-task control subset, specification completeness explains most of the gain. Targeted skills improve Codex reward by \resultnum{+0.136 [0.085, 0.191]} on 28 tasks (\resultnum{26/28} improving), and the corresponding Claude difference is \resultnum{+0.120 [0.067, 0.177]} (\resultnum{21/28}), but only \resultnum{25/280} with-skill Codex runs pass all strict diagnostics and key F1 moves only \resultnum{+0.003}.
\end{enumerate}

\section{Related work}

\mbench{} builds on credentialed critical-care databases, \mimic{}~\citep{johnson2023mimiciv} and \eicu{}~\citep{pollard2018eicu}, on community reference implementations such as MIT-LCP MIMIC-Code~\citep{johnson2018mimiccode}, and on the M3 agent infrastructure for clinical EHR access~\citep{attrach2025conversational}.
Prior clinical EHR benchmarks and agent methods cover few-shot prediction, text-to-SQL, structured-table question answering, cohort and feature extraction, or tool-using clinical agents~\citep{wornow2023ehrshot,lee2022ehrsql,shi2024ehragent,lee2025emragent,ehrstruct2025,fhiragentbench2025,jiang2025medagentbench}. Retrieval-augmented medical question answering is benchmarked separately by MIRAGE~\citep{xiong2024medrag}. Unlike RWE-bench's end-to-end observational-study protocols over \mimic{}~\citep{li2026rwebench}, \mbench{} isolates reusable derivation components such as SOFA, suspected infection, ventilation, vasopressor-equivalent dose, and rolling urine output, then scores reproductions against declared references under a direct \noskill{} versus \withskill{} ablation.

\mbench{} also differs from medical-calculator benchmarks, which test formula recall, extraction, or arithmetic from notes or synthetic patient descriptions~\citep{khandekar2024medcalcbench,medcalcAudit2026}. Real EHR derivations also depend on item identifiers, time windows, unit conversion, missingness conventions, and output grain. Harmonization efforts such as CLIF and OHDSI/OMOP standardize clinical data and definitions across settings~\citep{rojascarroll2024clif,hripcsak2015ohdsi}. \mbench{} instead asks whether coding agents can reproduce a fixed set of declared derivations under controlled conditions.

\emph{Agent skills} are reusable file-based procedural-knowledge artifacts that agents can discover and load at run time~\citep{anthropic2025agentskills}. SkillsBench studies such artifacts across domains, but only a few of its tasks involve clinical data, too few to establish whether skill effects are broad or reliable across EHR derivations~\citep{li2026skillsbench}. Structured-prompting and prompt-sensitivity work likewise show that procedural context can change model behavior~\citep{khattab2024dspy,zhuo2024prosa}. \mbench{} measures the effect of supplying task-specific procedural knowledge in a released, auditable form, without claiming a unique benefit for one packaging format.

Finally, \mbench{} inherits concerns from coding-agent benchmarks such as SWE-bench~\citep{jimenez2024swebench}: isolation, reproducibility, and contamination matter when agents can inspect code and files. Contamination is a known threat to such benchmarks and has motivated continuously updated or contamination-resistant evaluations~\citep{ravaut2025contaminationsurvey,wu2024antileak,zhang2025swebenchlive}. \mbench{} therefore reports OS-level isolation, artifact retention, and contamination checks, and treats schema perturbations as probes of schema familiarity only.

\section{Benchmark}
\label{sec:benchmark}

\paragraph{Benchmark at a glance.}
\mbench{} releases 28 task specifications across 15 clinical task families over \mimic{}~\citep{johnson2023mimiciv} and \eicu{}~\citep{pollard2018eicu}; clinician-reviewed procedural skills; the Docker isolation harness, evaluator, and contamination-audit code; and sanitized per-run evidence for the Codex campaigns, the submitted Claude sentinel campaign, and a full-coverage Claude replication. Metadata and hash manifests accompany the release. Derived task databases are not redistributed, while task specifications, evaluator code, and skills use a permissive open-source license. Asset terms appear in Table~\ref{tab:asset_licenses}.

\mbench{} contains 26 \mimic{} tasks and two \eicu{} tasks spanning severity scores, organ dysfunction, comorbidity, infection, respiratory support, medications, and rolling windows. Figure~\ref{fig:pipeline} summarizes the pipeline, and Table~\ref{tab:task_inventory} lists the family-level inventory.

\paragraph{Task format.}
Each task gives the agent a clinical derivation target and a task database. The agent must write \texttt{output.csv} with required identifier columns and value columns. Evaluation checks that the output is valid CSV, contains required columns, covers the expected key set, and matches reference value columns within task-specific tolerances. Appendix~\ref{sec:appendix_worked_example} traces one task, suspected infection, from clinical definition through output rows to diagnostics.

\paragraph{Standard and raw modes.}
Standard tasks remove reference target tables while leaving selected intermediate feature tables available. Raw tasks remove the target concept and, for \mimic{} tasks, remove the \texttt{mimiciv\_derived} schema so the agent must work from source clinical tables. This distinction tests whether skills matter more when implementation details such as item identifiers, source tables, unit conversions, and temporal joins cannot be bypassed through derived features.

\paragraph{Retrospective vs time-local framing.}
Several reference implementations adapted from public concept code use information available only after a real-time decision point (e.g., baseline creatinine uses discharge ICD codes for chronic-kidney-disease patients). \mbench{} is therefore a benchmark of \emph{retrospective concept derivations}, not real-time clinical prediction.

\paragraph{Conditions.}
The primary comparison is \noskill{} versus \withskill{} on the native schema. Table~\ref{tab:conditions} summarizes the main benchmark conditions and mechanism controls. The \noskill{} condition is a schema-plus-task baseline, not a deployed-system baseline such as retrieval over public concept code. The operational-spec control therefore measures how much of the effect comes from specification completeness rather than the skill format itself.

\begin{table}[t]
\centering
\caption{Glossary of benchmark conditions, controls, and metrics.}
\label{tab:conditions}
\footnotesize
\begin{tabular}{@{}p{0.22\linewidth}p{0.34\linewidth}p{0.34\linewidth}@{}}
\toprule
Term & What the agent receives & Purpose \\
\midrule
\noskill{} & Task instruction and task database & Schema-plus-task baseline; no curated derivation procedure \\
\withskill{} & \noskill{} inputs plus a task-specific procedural skill & Primary test of targeted procedural specification \\
Decoy skill & \noskill{} inputs plus a task-mismatched skill artifact & Controls for generic extra context and skill-file presence \\
Raw reference SQL & \noskill{} inputs plus reference SQL text & Tests reference-code structure; raw-task cells require SQL adaptation rather than answer copying \\
Operational spec & \noskill{} inputs plus complete non-SQL clinical rules, excluding item identifiers and dataset recipes & Separates rule completeness from dataset navigation and skill packaging \\
Schema perturbations & \noskill{} or \withskill{} inputs on obfuscated or restructured schemas & Probes schema-name and table-organization familiarity \\
\midrule
Metric & Definition & Interpretation \\
\midrule
Reward & Mean match rate over scored value columns, with missing, extra, and duplicate keys in each denominator & Partial agreement with the declared reference \\
Strict pass & Every evaluator check passes: required columns, key coverage, and a match-rate threshold on every scored column & Reliable end-to-end reproduction of the reference \\
Key F1 & Harmonic mean of key precision and reference-key recall & Output-grain agreement, separate from values \\
\bottomrule
\end{tabular}
\end{table}

\paragraph{Skill artifact.}
Each task-specific skill is a markdown procedural artifact in the spirit of Agent Skills~\citep{anthropic2025agentskills}. Skills are derived from public clinical definitions, MIT-LCP MIMIC-Code, and eICU concept-code conventions~\citep{johnson2018mimiccode,pollard2018eicu}, then adapted into implementation guidance for one declared reference convention, not universal clinical knowledge. Each task-specific skill and paired task specification was independently reviewed by two clinicians. Disagreements and requests for clarification were resolved through consensus discussion, and reviewer identifiers C1--C4 are reported in the provenance appendix. The four reviewing clinicians are co-authors of this paper, so the review is internal to the author team and is not external validation. This review checks definition provenance, task framing, and artifact plausibility. Skills may include formulas, item identifiers, table families, time-window rules, missingness conventions, unit conversions, output-grain reminders, and edge-case cautions. The primary skills contain no executable SQL snippets, hidden reference-output paths, row-level answers, exact result tables, or full solution programs. Raw SQL appears only in a matched-content control.

\paragraph{Reference outputs and metric.}
Reference outputs are adapted from public MIT-LCP concepts, eICU concept code, and benchmark-specific task semantics. The scalar reward is the mean match rate over scored value columns, not the pass fraction over tests. For task \(t\), scored columns \(C_t\), reference key set \(K_t\), and agent output \(A_t\), the evaluator left-joins agent rows to reference rows on the configured key columns, drops duplicate agent keys after counting them as penalties, and counts missing reference rows, duplicate agent keys, and extra agent keys against each column's denominator. The reward is
\[
R_t = \frac{1}{|C_t|}\sum_{c \in C_t}
\frac{\#\{k \in K_t: A_t(k,c) \approx G_t(k,c)\}}{|K_t| + \#\mathrm{extra\ keys} + \#\mathrm{duplicate\ keys}},
\]
where \(G_t\) denotes the hidden reference output table and \(\approx\) denotes exact equality or a task-specific numeric/timestamp tolerance. The evaluator also reports key precision, extra keys, row coverage, and pass/fail diagnostics. To separate row-grain errors from near-miss value errors, we additionally summarize key F1 and normalized MAE over ordered value columns. Unordered categorical columns remain exact-match only.

The MIT-LCP and eICU conventions used for these reference outputs are widely adopted research implementations, not clinically validated gold standards. Clinician review identified cases where public concept code diverges from source clinical definitions or omits clinically relevant edge cases. Appendix Table~\ref{tab:clinical_reference_audit} summarizes the highest-impact reference-validity caveats. We retain the public reference implementations for the current evidence package because changing them would redefine the benchmark target and require rerunning affected tasks. Future versions should add clinician-adjudicated reference targets alongside public-code targets, so that \mbench{} can separately measure agreement with community convention and agreement with clinical intent.

\paragraph{Dataset scale.}
Benchmark task databases are per-task DuckDB clones derived from credentialed MIMIC-IV and eICU data. MIMIC-IV tasks draw from the MIMIC-IV v3.1 release, and eICU tasks draw from the eICU Collaborative Research Database v2.0. Reference-output row counts range from \resultnum{41,296} rows for the Sepsis-3 cohort task to \resultnum{4,126,485} rows for rolling urine-output windows, with several interval and event tasks above \resultnum{700,000} rows. Per-task reference row counts are reported in Table~\ref{tab:task_details} in the appendix.

\begin{table}[t]
\centering
\caption{\mbench{} task inventory. Standard tasks retain selected intermediate features; raw tasks require reconstruction from source tables.}
\label{tab:task_inventory}
\begin{tabular}{lrrrr}
\toprule
Family & Tasks & Standard & Raw & Database \\
\midrule
APS III & 2 & 1 & 1 & MIMIC-IV \\
Charlson & 1 & 0 & 1 & MIMIC-IV \\
Baseline creatinine & 2 & 1 & 1 & MIMIC-IV \\
GCS & 2 & 0 & 2 & MIMIC-IV/eICU \\
KDIGO AKI & 2 & 1 & 1 & MIMIC-IV \\
MELD & 2 & 1 & 1 & MIMIC-IV \\
OASIS & 3 & 1 & 2 & MIMIC-IV/eICU \\
SAPS-II & 2 & 1 & 1 & MIMIC-IV \\
Sepsis-3 & 1 & 0 & 1 & MIMIC-IV \\
SIRS & 1 & 0 & 1 & MIMIC-IV \\
SOFA & 2 & 1 & 1 & MIMIC-IV \\
Suspected infection & 2 & 1 & 1 & MIMIC-IV \\
Urine-output rate & 2 & 1 & 1 & MIMIC-IV \\
Vasopressor equivalents & 2 & 1 & 1 & MIMIC-IV \\
Ventilation & 2 & 1 & 1 & MIMIC-IV \\
\midrule
Total & 28 & 11 & 17 & MIMIC-IV/eICU \\
\bottomrule
\end{tabular}

\end{table}

\paragraph{Data access and ethics.}
MIMIC-IV and eICU require credentialed access through PhysioNet. Benchmark task databases are derived from this credentialed data and are not redistributed. Researchers using \mbench{} must independently obtain PhysioNet credentials for the relevant source databases and rebuild the task databases with the released setup scripts (Appendix~\ref{sec:appendix_release}).

\paragraph{Data and code release.}
The source release (tag \texttt{\releasetag}, \url{\releaseurl}) contains the evaluation code, tasks, skills, controls, isolation harness, tests, run manifests, and table scripts. Two open evidence packages hold \texttt{result.json}, sanitized traces, and instructions for 1,960 runs (the 1,850 paper-facing runs, with egress logs, and 110 exploratory OSS runs, which have none), with source-to-public hash lineage and Croissant metadata~\citep{akhtar2024croissant}. Row-level agent outputs and tool outputs are withheld because they derive from credentialed data (Appendix~\ref{sec:appendix_release}).

\section{Experimental protocol}

\paragraph{Primary estimand.}
The primary estimand is the task-paired change in mean per-column match rate between \noskill{} and \withskill{} on native-schema tasks. Let \(R_{t,c,s,m,k}\) denote the reward defined in Section~\ref{sec:benchmark} for task \(t\), condition \(c\), schema \(s\), model \(m\), and trial \(k\). For native-schema tasks the per-task skill effect is
\[
\Delta_{t,m} = \mathbb{E}_k\!\left[R_{t,\,\withskill,\,\text{native},\,m,\,k}\right] - \mathbb{E}_k\!\left[R_{t,\,\noskill,\,\text{native},\,m,\,k}\right].
\]
The headline aggregate averages \(\Delta_{t,m}\) first over models within an arm and then over tasks, so each task contributes equally regardless of trial count or output size.
This estimand is identified within a single model provider: the two Codex models share a provider family. The Claude replication applies the same estimand to a second provider but, because of the harness differences described below, is a descriptive cross-provider check rather than evidence of a provider-general effect.

\paragraph{Inference and uncertainty.}
The campaign matrix is tiered and partly pilot-informed, so the headline estimand is task-balanced rather than run-weighted. We report non-parametric uncertainty over tasks using a task-cluster bootstrap for the model-averaged \(\Delta_t\), a family-cluster sensitivity analysis, a family-balanced bootstrap, and a Monte Carlo sign-flip test against zero. Bracketed intervals are 95\% non-parametric percentile bootstrap intervals over task-level deltas, not standard deviations or standard errors over repeated runs. These summaries are conditional on the task set.

\paragraph{Execution.}
Paper-quality runs execute inside Docker. The final primary Codex analysis uses \gptmini{} (\texttt{gpt-5.4-mini}) and \gptfull{} (\texttt{gpt-5.5}) over all 28 native tasks with five trials per cell plus matched-context controls, for \resultnum{760 canonical paper-facing runs}. The submitted Claude sentinel arm used Opus and Sonnet over eight paired tasks plus limited schema probes (\resultnum{190 runs}) and is retained as historical evidence. Each run records the task, condition, schema, agent, model, trial, timestamps, result hashes, token usage when available, and evaluator diagnostics. Release roots, rerun replacements, artifact counts, and compute are detailed in Appendices~\ref{sec:appendix_benchmark_details} and~\ref{sec:appendix_results_details}.

\paragraph{Full-coverage Claude replication.}
After submission we ran Claude Code with requested Opus (\texttt{claude-opus-4-7}) and Sonnet (\texttt{claude-sonnet-4-6}) on all 28 native tasks, \noskill{} and \withskill{}, five trials per cell, under the frozen task, skill, and evaluator definitions and the same Docker path: \resultnum{560 canonical runs} from three results roots on two machines, selected by a declared manifest, with \resultnum{656} failed, quarantined, superseded, or overlapping attempts inventoried and excluded (Appendix~\ref{sec:appendix_claude28}). The harness was not uniform: Claude Code 2.1.119 and 2.1.220 both occur, partly by condition (\resultnum{167/280} \noskill{} versus \resultnum{100/280} \withskill{} runs on 2.1.119). Model-slot overrides are recorded for \resultnum{230} runs and the rest use native routing, whose traces report ancillary \texttt{claude-haiku-4-5} usage in \resultnum{330} runs and Haiku assistant turns in \resultnum{5}. We therefore treat the 28-task contrast as descriptive and add a sensitivity analysis on tasks whose \noskill{} and \withskill{} cells share one recorded CLI version and routing profile per model. This matches recorded indicators, not every ancillary model call.

\paragraph{Isolation and contamination auditing.}
Paper campaigns enforce benchmark-internal isolation at the OS/container layer and fail closed if an agent can read hidden reference outputs, task internals, previous results, or evaluator internals. Post-run auditing excludes or fails traces that mention answer paths, evaluator internals, exact answer-copying behavior, or byte-identical outputs against decompressed answer files. The final Codex campaign passed publishability, isolation, filesystem-canary, and contamination-lint checks for all 760 canonical runs, as did all 560 canonical Claude replication runs, whose egress logs show no disallowed hosts or denied requests. Full audit counts appear in Appendices~\ref{sec:appendix_results_details} and~\ref{sec:appendix_claude28}.

\paragraph{Schema perturbations and matched-context controls.}
Schema perturbations run agents on obfuscated or restructured schemas while preserving underlying data values. They are schema-familiarity probes, not full data-value decontamination tests. Matched-context controls use decoy skills and raw reference SQL to test whether gains are caused merely by longer context, by direct reference-code exposure, or by task-specific procedural guidance. For raw tasks, raw reference SQL can reference intermediate tables removed from the task database, so these cells are SQL-adaptation controls, not executable answer-copying controls. The May 6 validity follow-up adds two non-primary controls: an operational-spec condition on an eight-task control subset for both Codex models, and transformed-schema \withskill{} cells for \gptmini{} over the 26 MIMIC tasks.

\section{Results}

\paragraph{Specification completeness accounts for most of the gain on the eight-task control subset.}
On the ten-task decoy/raw-SQL subset, decoy skills match \noskill{} (\resultnum{0.554} vs \resultnum{0.553}), arguing against generic extra context as the explanation. Raw reference SQL nearly matches targeted skills (\resultnum{0.837} vs \resultnum{0.833}, \(\Delta=\)\resultnum{0.003}), so reference-implementation structure already carries most of the signal once it is present (Appendix Table~\ref{tab:codex_matched_controls}). Operational specifications (complete non-SQL derivation rules without item identifiers or dataset recipes) improve over \noskill{} by \resultnum{+0.162 [0.072, 0.260]} on the eight-task subset, while clinician-reviewed targeted skills add only \resultnum{+0.024 [-0.034, 0.099]} on top (Table~\ref{tab:codex_operational_spec_control}). The combined targeted-skill effect on all 28 native tasks is \resultnum{+0.136 [0.085, 0.191]} (Table~\ref{tab:codex_native_skill_effect}), family-cluster sensitivity \resultnum{[0.065, 0.207]}, family-balanced bootstrap \resultnum{+0.120 [0.057, 0.190]} (Appendix Table~\ref{tab:codex_family_sensitivity}), sign-flip \(p<10^{-4}\), with \resultnum{26/28 tasks} improving and ordered-value nMAE falling from \resultnum{0.116} to \resultnum{0.060}. Skills are therefore one clinician-reviewable delivery format for explicit procedural specification. The control subsets were purposive: both task lists were fixed in the benchmark matrix before their first runs, and they over-represent large-effect tasks (the targeted-skill delta is \resultnum{+0.280} on the 10 decoy/raw-SQL tasks versus \resultnum{+0.056} on the other 18, and \resultnum{+0.186} on the eight operational-spec tasks versus \resultnum{+0.116}), and the operational-spec subset has no \eicu{} task. Mechanism conclusions apply to these subsets (Appendix Table~\ref{tab:codex_control_subsets}).

\begin{table}[t]
\centering
\caption{Native-schema Codex skill effect and diagnostics. Reward is mean value-column match rate; key F1 measures row grain; ordered nMAE is lower-is-better. Intervals are 95\% task-bootstrap intervals.}
\label{tab:codex_native_skill_effect}
\resizebox{\textwidth}{!}{\begin{tabular}{lrrrr}
\toprule
Measure & No skill & With skill & Delta & Improved tasks \\
\midrule
Reward / GPT 5.4 Mini & 0.639 & 0.802 & 0.162 [0.099, 0.232] & 26/28 \\
Reward / GPT 5.5 & 0.767 & 0.876 & 0.109 [0.060, 0.166] & 24/28 \\
Reward / task-balanced & 0.703 & 0.839 & 0.136 [0.085, 0.191] & 26/28 \\
Key F1 / task-balanced & 0.962 & 0.965 & 0.003 [-0.010, 0.018] & 5/28 \\
Ordered nMAE / task-balanced & 0.116 & 0.060 & -0.056 [-0.085, -0.029] & 23/26 \\
\bottomrule
\end{tabular}
}
\end{table}

\paragraph{Few runs pass strictly, and skills barely move row grain.}
The continuous reward measures partial agreement on matched keys, not end-to-end correctness. Only \resultnum{87/760} runs pass every strict diagnostic (\resultnum{13/280} \noskill{} and \resultnum{25/280} \withskill{} in the primary comparison; the other \resultnum{49} are raw-SQL control runs, \resultnum{49/100} against \resultnum{7/100} targeted-skill runs on the same ten tasks, although mean reward is nearly equal). Most strict failures are value errors on correctly keyed rows: \resultnum{468} of the \resultnum{522} failing primary-comparison runs meet the key-coverage threshold and miss only a value-match threshold (0.85 to 1.0 per task, exact on 10 of 11 standard tasks). Skills barely move row grain: task-balanced key F1 changes by only \resultnum{+0.003} and improves on only \resultnum{5/28} tasks (it falls on \resultnum{3}), even though many of the targeted dataset-navigation cautions concern intervals and grain. The headline gain is therefore concentrated in numeric value-column agreement on rows where the agent already produced a matching key.

\begin{table}[t]
\centering
\caption{Codex operational-spec follow-up on the eight-task control subset. Values are task-balanced after averaging the two Codex models within each task.}
\label{tab:codex_operational_spec_control}
\resizebox{\textwidth}{!}{\begin{tabular}{lrrrrr}
\toprule
Comparison & Tasks & No skill & Operational spec & Targeted skill & Delta \\
\midrule
Operational spec--no skill & 8 & 0.606 & 0.767 & -- & 0.162 [0.072, 0.260] \\
Targeted skill--operational spec & 8 & -- & 0.767 & 0.792 & 0.024 [-0.034, 0.099] \\
\bottomrule
\end{tabular}
}
\end{table}

\paragraph{Heterogeneity and failure modes.}
The largest gains occur on tasks requiring implementation-level knowledge: rolling urine-output windows, MELD edge cases, suspected-infection timing, baseline-creatinine hierarchy, and ventilation status. KDIGO, Charlson, standard vasopressor-equivalent, SOFA, and several already-high-baseline scores remain near zero or negative. All eight flattest tasks (Appendix Table~\ref{tab:codex_skill_declines}) keep key F1 at \resultnum{1.000} in both conditions: six are near ceiling without skills (no-skill reward at least \resultnum{0.94}), and raw KDIGO and standard APS~III keep value errors that skills do not fix. Grain errors concentrate in raw Sepsis-3, raw ventilation, and raw suspected infection (key F1 \resultnum{0.68} to \resultnum{0.87}). Per-task deltas and mode splits appear in Appendix~\ref{sec:appendix_results_details}.

\paragraph{Schema perturbations: descriptive robustness.}
Schema obfuscation and restructuring alter the database interface while preserving underlying values. In a \gptmini{} follow-up over the 26 MIMIC tasks, transformed-schema \withskill{} cells remain positive when matched to the transformed-schema \noskill{} cells from the earlier probe: \resultnum{+0.135 [0.068, 0.208]} on obfuscated schemas and \resultnum{+0.110 [0.057, 0.170]} on restructured schemas, or \resultnum{0.862} and \resultnum{0.704} of the native-schema gain (Appendix Table~\ref{tab:codex_schema_skill_generalization}). We read these ratios descriptively. Treating the shortfall as the share of the effect due to schema-name memorization would require that the perturbations change nothing else about task difficulty and that the April no-skill probe and May with-skill runs are exchangeable. Neither is established, and the transformed-schema no-skill runs are not in the retained evidence packages (Appendix~\ref{sec:appendix_release}). The probe says nothing about value-level memorization or provider, hospital, or database generalization.

\paragraph{Full-coverage Claude replication.}
Across all 28 tasks, Claude reward is \resultnum{0.747} without and \resultnum{0.867} with targeted skills, a task-balanced difference of \resultnum{+0.120 [0.067, 0.177]} with \resultnum{21/28} tasks positive (Table~\ref{tab:claude28_sensitivity}). The Opus difference is \resultnum{+0.083 [0.038, 0.135]} (\resultnum{25/28}) and the Sonnet difference \resultnum{+0.156 [0.087, 0.231]} (\resultnum{20/28}; Appendix Table~\ref{tab:claude28_native_skill_effect}). Restricting to the \resultnum{19} tasks with matched recorded CLI version and routing profile gives \resultnum{+0.119 [0.054, 0.192]} (\resultnum{14/19}), so the positive average does not depend on the tasks with recorded harness imbalance. It does not show that CLI or routing changes had no effect. As with Codex, the difference is mostly value agreement: strict passes rise from \resultnum{25/280} to \resultnum{37/280} and key F1 by \resultnum{+0.008}. Differences concentrate on the same implementation-heavy families (urine-output windows, baseline creatinine, MELD, suspected infection), while SOFA, APS~III, SAPS~II, KDIGO, and Charlson stay near zero (Appendix Table~\ref{tab:claude28_task_deltas}). The replication supersedes the submitted eight-task Claude sentinel (\resultnum{+0.166 [0.085, 0.259]}), which is retained as historical evidence. The lower full-coverage estimate is consistent with that purposive subset over-representing large-effect tasks. This comparison is not a provider ranking.

\begin{table}[t]
\centering
\caption{Full-coverage Claude replication: task-balanced native-schema \withskill{} minus \noskill{} reward on all 28 tasks, on the 20 tasks whose \noskill{} and \withskill{} cells share one recorded Claude Code version (CLI) per model, and on the 19 that also share one routing profile. Intervals are 95\% task bootstraps. Strict passes and key F1 are separate from reward.}
\label{tab:claude28_sensitivity}
\resizebox{\textwidth}{!}{\begin{tabular}{lrrrrrr}
\toprule
Task set & Tasks & Runs & Delta [95\% CI] & Positive & Strict passes (no/with) & Key F1 delta \\
\midrule
All tasks & 28 & 560 & 0.120 [0.067, 0.177] & 21/28 & 25/280 vs 37/280 & 0.008 \\
CLI-version matched & 20 & 400 & 0.113 [0.050, 0.185] & 15/20 & 19/200 vs 27/200 & 0.008 \\
CLI and routing matched & 19 & 380 & 0.119 [0.054, 0.192] & 14/19 & 19/190 vs 27/190 & 0.008 \\
\bottomrule
\end{tabular}
}
\end{table}

\paragraph{Capability floor.}
An exploratory local \texttt{gpt-oss-20b} arm is non-publishable under our release criteria, but its near-zero task-paired delta of \resultnum{+0.008 [-0.072, 0.088]} suggests that procedural specifications do not substitute for base coding capability (Appendix~\ref{sec:appendix_results_details}).

\section{Discussion and limitations}

\paragraph{What \mbench{} measures.}
\mbench{} measures whether agents can implement declared clinical EHR reference derivations and whether explicit procedural specifications close implementation gaps. The tasks cover critical-care severity scores, organ dysfunction, comorbidity, infection, respiratory support, medications, and rolling windows over \mimic{} and \eicu{}. The benchmark does not measure bedside reasoning, diagnosis, treatment planning, real-time prediction, or downstream study validity.

\paragraph{Procedural specification as an auditable artifact.}
Relative to EHR text-to-SQL, structured-QA, and interactive-agent benchmarks~\citep{lee2022ehrsql,ehrstruct2025,jiang2025medagentbench}, \mbench{} evaluates the reproduction of declared clinical derivations instead of answers to EHR questions. It turns procedural knowledge such as item identifiers, time windows, and tie-breaking rules into clinician-reviewed, versioned artifacts. These artifacts help some derivations and leave others flat or negative when grain, timing, or reconstruction fail.

\paragraph{Interpretation limits and what \mbench{} does not claim.}
The estimand identifies reference agreement, not clinical correctness, hidden-knowledge transfer, or decontaminated reasoning. The evidence is a Codex headline plus a descriptive Claude replication with disclosed harness differences, so it is not provider-general. Public concept code likely appears in pretraining, and the schema probes address schema familiarity only, so the results are not memorization-free. Operational specifications without item identifiers already capture most of the gain, so the effect is not unique to the skill format. Skills barely change key F1 on average and improve output grain on only a few tasks. With 26 \mimic{} and two \eicu{} tasks, the evidence is not dataset- or hospital-general. The small targeted-skill advantage over operational specifications is suggestive of dataset-navigation value, not a powered unique-skill effect.

\paragraph{Statistical and external validity.}
The campaign is pilot-informed, not a balanced confirmatory design. The Claude replication matches Codex in task coverage and isolation but not in harness uniformity. A causal skill contrast for Claude needs a declared rerun of the affected cells under one frozen CLI version and routing policy. Broader \eicu{} and cross-site coverage is the main external-validity priority.

\paragraph{Guidance for agent builders and practitioners.}
Given the residual failures, agent builders should pair procedural guidance with executable checks. Before coding, declare the output grain, keys, window boundaries, units, missingness and tie-breaking rules, and source mappings. Before accepting output, validate columns, key coverage, duplicate and extra keys, cohort size, value ranges, intermediate counts, and boundary cases, because skills barely move key F1 or strict pass. Agents also need reliable primitives for temporal joins, unit normalization, and item-code mapping, and should expose intermediate cohorts and values for inspection. Contestable choices should be surfaced, not silently resolved, and practitioners should treat agent output as an untrusted draft that needs independent clinical review before consequential use. Expert-curated examples help most when they carry the definition, dataset mappings, expected intermediate and final outputs, edge cases, tests, provenance, and known limitations, not only final SQL.

\paragraph{Safety and misuse.}
\mbench{} evaluates derived variables for supervised research workflows, not autonomous clinical decision-making. Incorrect derivations propagate into cohorts, risk adjustment, and downstream conclusions, and high reward can coexist with consequential grain, timing, or missingness errors. References can embed contestable choices (Appendix Table~\ref{tab:clinical_reference_audit}), so human audit and provenance tracking remain necessary.

\section{Conclusion}

\mbench{} is an evaluation benchmark for procedural specification in clinical EHR derivation agents. Explicit procedural specifications, instantiated here as clinician-reviewed skills, improve agreement with declared reference outputs for Codex, with a directionally consistent descriptive Claude replication, while leaving important failures in temporal joins, output grain, and raw intermediate reconstruction. The result shows that schema exploration alone is often insufficient and gives readers an artifact-backed way to inspect residual errors requiring human audit.

\mbench{} is designed to be extended through \texttt{task.toml} files with paired reference-output implementations, evaluation configuration, and clinician-reviewed skills. Researchers can add tasks, skills, controls, and agents under the same \noskill{} versus \withskill{} comparison, isolation harness, and evidence standards to measure the value of agent skills in other clinical EHR settings. Future versions should add mechanism controls on all 28 tasks, clinician-adjudicated targets, multiple accepted derivations, broader \eicu{} and cross-site tasks, and care-process task families.

\section*{Acknowledgments and Disclosure of Funding}
\textbf{Funding.} This project was not funded by any grant. LAC is funded by the National Institutes of Health through DS-I Africa U54 TW012043-01 and Bridge2AI OT2OD032701, the National Science Foundation through ITEST \#2148451, a grant of the Boston-Korea Innovative Research Project (RS-2024-00403047), and a grant of the Korea Health Technology R\&D Project (RS-2024-00439677) through the Korea Health Industry Development Institute (KHIDI), funded by the Ministry of Health \& Welfare, Republic of Korea.

\textbf{Competing interests.} The authors declare no competing interests.

\bibliographystyle{plainnat}
\bibliography{bib/references}

\clearpage
\appendix
\section{Benchmark details}
\label{sec:appendix_benchmark_details}
\suppressfloats[t]

\paragraph{Task families.}
The 15 task families cover severity scores, neurological status, renal and liver dysfunction, comorbidity, infection, respiratory support, medications, and rolling-window data engineering. Ten families derive from published scores and definitions: SOFA~\citep{vincent1996sofa}, APS~III~\citep{knaus1991apacheiii}, SAPS~II~\citep{legall1993sapsii}, OASIS~\citep{johnson2013oasis}, GCS~\citep{teasdale1974gcs}, the Charlson comorbidity index~\citep{charlson1987comorbidity}, SIRS~\citep{bone1992sirs}, MELD~\citep{kamath2001meld}, KDIGO AKI staging~\citep{kdigo2012aki}, and Sepsis-3~\citep{singer2016sepsis3}. The reference SQL adapts each definition to its database following the source basis in Table~\ref{tab:task_review_provenance}. The full task inventory is generated from \texttt{benchmark/tasks/*/*/task.toml}.

\paragraph{Required output.}
Each task specifies required identifier columns and value columns. The evaluator computes row coverage, required-column validity, key precision/F1, value-column match rates, and ordered-column normalized MAE. The reported reward is the mean value-column match rate. Ordered-column nMAE is computed on matched keys, normalizes absolute error by the observed reference range for each scored column, and clips per-value normalized errors at one.

\begin{table}[t]
\centering
\caption{Per-task evaluation details. Reference rows are reference-output rows before task-specific evaluation joins.}
\label{tab:task_details}
\resizebox{\textwidth}{!}{\begin{tabular}{llllrr}
\toprule
Task & Mode & Database & Key grain & Scored columns & Reference rows \\
\midrule
eicu-gcs & raw & eICU & stay & 4 & 200859 \\
eicu-oasis & raw & eICU & stay & 11 & 200859 \\
mimic-apsiii-24h & standard & MIMIC-IV & stay & 17 & 94458 \\
mimic-apsiii-24h-raw & raw & MIMIC-IV & stay & 17 & 94458 \\
mimic-charlson-raw & raw & MIMIC-IV & admission & 19 & 546028 \\
mimic-creatinine-baseline & standard & MIMIC-IV & admission & 3 & 546028 \\
mimic-creatinine-baseline-raw & raw & MIMIC-IV & admission & 3 & 546028 \\
mimic-gcs-24h-raw & raw & MIMIC-IV & stay & 4 & 94458 \\
mimic-kdigo-48h & standard & MIMIC-IV & stay & 4 & 94458 \\
mimic-kdigo-48h-raw & raw & MIMIC-IV & stay & 4 & 94458 \\
mimic-meld-24h & standard & MIMIC-IV & stay & 1 & 94458 \\
mimic-meld-24h-raw & raw & MIMIC-IV & stay & 1 & 94458 \\
mimic-oasis-24h & standard & MIMIC-IV & stay & 11 & 94458 \\
mimic-oasis-24h-raw & raw & MIMIC-IV & stay & 11 & 94458 \\
mimic-sapsii-24h & standard & MIMIC-IV & stay & 16 & 94458 \\
mimic-sapsii-24h-raw & raw & MIMIC-IV & stay & 16 & 94458 \\
mimic-sepsis3-raw & raw & MIMIC-IV & stay & 8 & 41296 \\
mimic-sirs-24h-raw & raw & MIMIC-IV & stay & 5 & 94458 \\
mimic-sofa-24h & standard & MIMIC-IV & stay & 7 & 94458 \\
mimic-sofa-24h-raw & raw & MIMIC-IV & stay & 7 & 94458 \\
mimic-suspicion-infection & standard & MIMIC-IV & event & 3 & 949901 \\
mimic-suspicion-infection-raw & raw & MIMIC-IV & event & 3 & 949901 \\
mimic-urine-output-rate & standard & MIMIC-IV & stay-window & 6 & 4126485 \\
mimic-urine-output-rate-raw & raw & MIMIC-IV & stay-window & 6 & 4126485 \\
mimic-vasopressor-equivalents & standard & MIMIC-IV & interval & 1 & 781067 \\
mimic-vasopressor-equivalents-raw & raw & MIMIC-IV & interval & 1 & 781067 \\
mimic-ventilation & standard & MIMIC-IV & interval & 1 & 144812 \\
mimic-ventilation-raw & raw & MIMIC-IV & interval & 1 & 144812 \\
\bottomrule
\end{tabular}
}
\end{table}

\begin{table}[t]
\centering
\caption{Task-level clinical review provenance. Each task is reviewed by two clinicians, identified as C1--C4, and every review covered the task definition, schema, and output. Reviewer identifiers are described in Table~\ref{tab:clinical_reviewer_ids}.}
\label{tab:task_review_provenance}
\begin{tabular}{lll}
\toprule
Task & Reviewers & Source basis \\
\midrule
eicu-gcs & C1/C4 & eICU concept conventions \\
eicu-oasis & C2/C4 & eICU concept conventions \\
mimic-apsiii-24h & C1/C4 & MIT-LCP concept conventions \\
mimic-apsiii-24h-raw & C1/C4 & MIT-LCP concept conventions \\
mimic-charlson-raw & C1/C4 & MIT-LCP concept conventions \\
mimic-creatinine-baseline & C1/C4 & Public clinical definitions \\
mimic-creatinine-baseline-raw & C1/C4 & Public clinical definitions \\
mimic-gcs-24h-raw & C1/C4 & MIT-LCP concept conventions \\
mimic-kdigo-48h & C1/C4 & Public clinical definitions \\
mimic-kdigo-48h-raw & C1/C4 & Public clinical definitions \\
mimic-meld-24h & C2/C4 & MIT-LCP concept conventions \\
mimic-meld-24h-raw & C2/C4 & MIT-LCP concept conventions \\
mimic-oasis-24h & C2/C4 & MIT-LCP concept conventions \\
mimic-oasis-24h-raw & C2/C4 & MIT-LCP concept conventions \\
mimic-sapsii-24h & C2/C4 & MIT-LCP concept conventions \\
mimic-sapsii-24h-raw & C2/C4 & MIT-LCP concept conventions \\
mimic-sepsis3-raw & C2/C4 & Sepsis-3/MIT-LCP conventions \\
mimic-sirs-24h-raw & C2/C4 & Public clinical definitions \\
mimic-sofa-24h & C3/C2 & MIT-LCP concept conventions \\
mimic-sofa-24h-raw & C3/C2 & MIT-LCP concept conventions \\
mimic-suspicion-infection & C3/C2 & MIT-LCP concept conventions \\
mimic-suspicion-infection-raw & C3/C2 & MIT-LCP concept conventions \\
mimic-urine-output-rate & C3/C2 & MIT-LCP concept conventions \\
mimic-urine-output-rate-raw & C3/C2 & MIT-LCP concept conventions \\
mimic-vasopressor-equivalents & C3/C2 & Public clinical definitions \\
mimic-vasopressor-equivalents-raw & C3/C2 & Public clinical definitions \\
mimic-ventilation & C3/C2 & MIT-LCP concept conventions \\
mimic-ventilation-raw & C3/C2 & MIT-LCP concept conventions \\
\bottomrule
\end{tabular}

\end{table}

\begin{table}[t]
\centering
\caption{Clinical reviewer identifiers. Descriptions summarize credentials relevant to clinical derivation review.}
\label{tab:clinical_reviewer_ids}
\begin{tabular}{lp{0.78\textwidth}}
\toprule
Reviewer ID & Description \\
\midrule
C1 & MD, MS; board-certified vascular and trauma surgeon; associate professor at a tertiary academic center; clinical registry and administrative-data researcher; MIMIC-IV/eICU experience. \\
C2 & MD, PhD; board-certified intensivist and emergency physician; clinical epidemiologist; ICU EHR researcher; MIMIC-IV/eICU experience. \\
C3 & MD; doctoral candidate and clinical AI researcher; expertise in explainable AI, ML-based clinical decision support, and multimodal prediction models; MIMIC-IV/eICU experience. \\
C4 & MD, PhD; associate professor at a medical school; physician and clinical AI researcher; expertise in ICU/EHR data research, multimodal clinical foundation models, and clinical AI evaluation; MIMIC-IV research experience. \\
\bottomrule
\end{tabular}

\end{table}

\begin{table}[t]
\centering
\caption{Reference-validity audit for selected task families. These caveats describe the declared benchmark target, not new exclusions from scoring.}
\label{tab:clinical_reference_audit}
\resizebox{\textwidth}{!}{\begin{tabular}{p{0.20\textwidth}p{0.34\textwidth}p{0.38\textwidth}}
\toprule
Family & Reference convention retained & Validity caveat \\
\midrule
Suspected infection / Sepsis-3 & Antibiotic--microbiology timing rule from public concept code & Broad culture inclusion can overcall suspected infection; downstream Sepsis-3 inherits this target. \\
SOFA / Sepsis-3 & Missing organ components default to zero & Agreement may reflect the public-code missingness convention rather than adjudicated organ failure. \\
APS III & MIMIC-Code component implementation & Known respiratory and oxygenation pathway discrepancies are retained for reproducibility. \\
Vasopressor equivalents & Five-agent norepinephrine-equivalent mapping & Angiotensin II administrations are not scored in the current reference. \\
eICU OASIS & eICU APACHE abstraction tables where source charting is not equivalent & eICU raw tasks are not schema-symmetric with MIMIC raw reconstruction. \\
\bottomrule
\end{tabular}
}
\end{table}

\paragraph{Retained private evidence.}
The submitted analysis retains \texttt{result.json}, \texttt{output.csv}, \texttt{instruction.md}, \texttt{trace.jsonl}, and \texttt{egress.jsonl} for every run in a private archive. The final Codex analysis uses 760 canonical runs from \texttt{release-20260502-codex-v11} plus \texttt{review-rerun-20260504-codex}. The retained Codex bundle has 800 directories because it also keeps 40 superseded early SQL-bearing skill runs for provenance, not analysis. The submitted Claude sentinel analysis combines \texttt{release-20260502-claude-provider} with \texttt{review-rerun-20260504-claude}. \texttt{artifact\_hash\_manifest.json} records SHA-256 hashes and byte sizes. Open packages derived from these archives are described in Appendix~\ref{sec:appendix_release}.

\begin{table}[t]
\centering
\caption{Retained paper-facing artifact directories after rerun replacement. Retained size is the uncompressed sum of listed files, not credentialed task databases. Codex has 760 canonical analysis runs plus 40 provenance-only directories.}
\label{tab:artifact_package}
\begin{tabular}{lrr}
\toprule
 & Codex primary & Claude sentinel \\
\midrule
Canonical runs & 760 & 190 \\
Provenance-only dirs & 40 & 0 \\
Retained dirs & 800 & 190 \\
result.json & 800 & 190 \\
output.csv & 800 & 190 \\
instruction.md & 800 & 190 \\
trace.jsonl & 800 & 190 \\
egress.jsonl & 800 & 190 \\
Retained size & 48.0 GB & 10.8 GB \\
\bottomrule
\end{tabular}

\end{table}

\begin{table}[t]
\centering
\caption{Compute and storage for the retained evidence package. Agent wall-clock sums the per-run agent execution time recorded in \texttt{result.json}. Selected dirs are retained paper-facing run directories on disk. Codex and Claude token rows reflect each provider's native usage reporting (Codex: total prompt tokens; Claude: split into uncached vs.\ cached input under provider prompt caching), so uncached counts are not directly comparable across arms.}
\label{tab:compute_resources}
\begin{tabular}{lrr}
\toprule
 & Codex primary & Claude sentinel \\
\midrule
Canonical runs & 760 & 190 \\
Provenance-only dirs & 40 & 0 \\
Retained dirs & 800 & 190 \\
Hardware & MacBook Pro (M3 Pro, 18 GB) & MacBook Pro (M3 Pro, 18 GB) \\
Agent wall-clock h & 56.4 & 17.5 \\
Uncached input & 586,293,569 & 16,657 \\
Cached input & 552,619,008 & 307,338,270 \\
Output tokens & 15,732,394 & 3,641,390 \\
Selected dirs & 48.1 GB & 10.8 GB \\
\bottomrule
\end{tabular}

\end{table}

\begin{table}[t]
\centering
\caption{Source assets, release terms, and access restrictions.}
\label{tab:asset_licenses}
\resizebox{\textwidth}{!}{\begin{tabular}{p{0.24\textwidth}p{0.24\textwidth}p{0.44\textwidth}}
\toprule
Asset & Role & Access, license, or terms \\
\midrule
M4Bench code, tasks, evaluator, skills & Benchmark artifact & MIT license (repository LICENSE); camera-ready source release tagged m4bench-neurips2026-camera-ready. \\
MIMIC-IV v3.1 & Source EHR data for 26 tasks & PhysioNet credentialed access and MIMIC-IV data-use terms (\url{https://physionet.org/content/mimiciv/3.1/}); derived databases are not openly redistributed. \\
eICU Collaborative Research Database v2.0 & Source EHR data for 2 tasks & PhysioNet credentialed access and eICU data-use terms (\url{https://physionet.org/content/eicu-crd/2.0/}); derived databases are not openly redistributed. \\
MIT-LCP MIMIC-Code / eICU concept conventions & Reference SQL lineage and procedural conventions & Credited and cited; adapted SQL/procedural conventions retained with provenance. \\
Per-run evidence packages & Evaluator results, sanitized traces, instructions, egress logs, hash lineage & Open packages withhold row-level agent outputs and tool outputs; originals stay private under source-data use terms. \\
\bottomrule
\end{tabular}
}
\end{table}

\paragraph{Fresh campaign checklist.}
Before a paper run, execute benchmark preflight, run an isolation canary as the same user and command path as the agent, create a fresh result root, and record the exact matrix summary. After the run, audit traces for benchmark-internal access and compare output hashes against answer files. Other provider-comparison campaigns should use separate fresh roots and remain exploratory unless promoted by the same artifact standard.

\section{Worked derivation example: suspected infection}
\label{sec:appendix_worked_example}

\paragraph{Clinical definition.}
Suspected infection pairs systemic antibiotic administration with culture collection, following the operationalization of \citet{seymour2016sepsis} used for Sepsis-3~\citep{singer2016sepsis3}: a culture up to 72 hours before or up to 24 hours after an antibiotic start.

\paragraph{Dataset mapping.}
In \mimic{}, antibiotics come from systemic \texttt{prescriptions} orders (topical, ophthalmic, and otic routes excluded) and cultures from \texttt{microbiologyevents}. A culture is positive when an organism is reported. Cultures with a collection date but no time use day-level windows (three days before, one day after). If both directions match, the culture-before-antibiotic match takes precedence, and within a direction the earliest eligible culture is chosen. The standard task keeps the curated antibiotic table, and the raw task removes it.

\paragraph{Output grain and scoring.}
The output has one row per antibiotic order keyed by \texttt{subject\_id} and \texttt{ab\_id}, a per-subject row number ordered by start time, stop time, and antibiotic name (\resultnum{949,901} reference rows). The reward averages match rates for \texttt{suspected\_infection}, \texttt{positive\_culture}, and \texttt{specimen}. A strict pass additionally requires every required column, full key coverage, and value agreement at the task threshold (exact for the standard task, 0.85 for the raw task).

\paragraph{What the diagnostics show.}
On the standard task in the primary Codex comparison, key precision is \resultnum{1.000} in both conditions and mean key F1 is at least \resultnum{0.999}, so both conditions get the grain right. Reward rises from \resultnum{0.720} to \resultnum{0.820}, and the unscored diagnostic match on \texttt{culture\_time} rises from \resultnum{0.46} to \resultnum{0.69}: the skill mainly corrects which culture is paired to each antibiotic (precedence and earliest-culture rules). No run in either condition passes strictly (\resultnum{0/10} each). On the raw task, where the agent must also rebuild the antibiotic table, grain is a problem as well: mean key F1 is \resultnum{0.725} without and \resultnum{0.865} with the skill. A clinically plausible rule applied with the wrong window or tie-break therefore still produces incorrect rows, and the reference itself inherits the broad culture inclusion noted in Table~\ref{tab:clinical_reference_audit}.

\section{Control subset selection}
\label{sec:appendix_control_subsets}

The decoy-skill and raw-SQL controls share one purposive 10-task list, committed to the benchmark matrix on 2 May 2026 (commit \texttt{3405cc45}) before the first control run on 3 May. It spans \resultnum{6/15} families, five standard and five raw tasks, and nine \mimic{} tasks plus one \eicu{} task. The operational-spec control uses the same eight tasks as the sentinel list that has been in the benchmark matrix since 23 April 2026 (\texttt{4245cc81}), twelve days before the first operational-spec run on 6 May. The commits that added the operational-spec files and the condition itself (\texttt{fce2bd24}, \texttt{949836d9}) were authored 62 and 70 minutes after that run started, so the pre-specification rests on the earlier list. The list spans \resultnum{8/15} families, three standard and five raw tasks, and no \eicu{} task. Commit times are author dates compared in UTC, and committer dates reflect a later history rewrite. The eight-task list was chosen from a single-seed Opus pilot with 27 valid tasks and holds five pilot high-delta tasks, one flat task, and two pilot-negative tasks (categories in \texttt{benchmark/matrix.py}). With a budget of 100 runs per control, we placed the decoy-skill and raw-SQL controls on the ten tasks with the largest skill delta in the earlier April 2026 Codex campaign (\texttt{codex-full-20260428}), where a difference between conditions was most likely to show. Under the Opus-pilot categories these are eight high-delta tasks, one flat task, and one negative task. Selection used the April runs, and the control results reported here come from separate May runs. Table~\ref{tab:codex_control_subsets} reports membership and the final targeted-skill delta per task. The subset deltas (\resultnum{+0.280} and \resultnum{+0.186}) exceed their complements (\resultnum{+0.056} and \resultnum{+0.116}), so control results describe these subsets rather than the full benchmark.

\begin{table}[t]
\centering
\caption{Control-subset membership and final Codex targeted-skill delta (task-balanced over both models) for all 28 tasks. Each task's family follows from its name, and families are listed in Table~\ref{tab:task_inventory}.}
\label{tab:codex_control_subsets}
\resizebox{\textwidth}{!}{\begin{tabular}{lllrcc}
\toprule
Task & Mode & Database & Delta & Decoy/raw SQL & Op. spec \\
\midrule
mimic-urine-output-rate-raw & raw & MIMIC-IV & 0.471 & yes & yes \\
mimic-urine-output-rate & standard & MIMIC-IV & 0.460 & yes &  \\
mimic-meld-24h & standard & MIMIC-IV & 0.354 & yes &  \\
mimic-oasis-24h & standard & MIMIC-IV & 0.296 &  & yes \\
mimic-suspicion-infection-raw & raw & MIMIC-IV & 0.291 & yes &  \\
mimic-meld-24h-raw & raw & MIMIC-IV & 0.283 & yes &  \\
eicu-oasis & raw & eICU & 0.272 & yes &  \\
mimic-ventilation & standard & MIMIC-IV & 0.246 & yes & yes \\
mimic-oasis-24h-raw & raw & MIMIC-IV & 0.221 &  &  \\
mimic-creatinine-baseline-raw & raw & MIMIC-IV & 0.180 & yes & yes \\
mimic-creatinine-baseline & standard & MIMIC-IV & 0.142 & yes &  \\
mimic-vasopressor-equivalents-raw & raw & MIMIC-IV & 0.123 &  & yes \\
mimic-suspicion-infection & standard & MIMIC-IV & 0.100 & yes & yes \\
mimic-apsiii-24h-raw & raw & MIMIC-IV & 0.055 &  &  \\
mimic-sapsii-24h & standard & MIMIC-IV & 0.055 &  &  \\
eicu-gcs & raw & eICU & 0.054 &  &  \\
mimic-ventilation-raw & raw & MIMIC-IV & 0.051 &  &  \\
mimic-sapsii-24h-raw & raw & MIMIC-IV & 0.038 &  &  \\
mimic-sepsis3-raw & raw & MIMIC-IV & 0.035 &  & yes \\
mimic-sofa-24h-raw & raw & MIMIC-IV & 0.034 &  & yes \\
mimic-apsiii-24h & standard & MIMIC-IV & 0.023 &  &  \\
mimic-gcs-24h-raw & raw & MIMIC-IV & 0.023 &  &  \\
mimic-sirs-24h-raw & raw & MIMIC-IV & 0.016 &  &  \\
mimic-sofa-24h & standard & MIMIC-IV & 0.010 &  &  \\
mimic-vasopressor-equivalents & standard & MIMIC-IV & 0.006 &  &  \\
mimic-charlson-raw & raw & MIMIC-IV & 0.001 &  &  \\
mimic-kdigo-48h & standard & MIMIC-IV & -0.002 &  &  \\
mimic-kdigo-48h-raw & raw & MIMIC-IV & -0.040 &  &  \\
\bottomrule
\end{tabular}
}
\end{table}

\section{Release, verification, and provenance}
\label{sec:appendix_release}

\paragraph{Source release.}
The tag \texttt{\releasetag} of \url{https://github.com/hannesill/m4} holds the benchmark code, skills, harness, evaluator, tests, run manifests, and table scripts as they stood after the last campaign. It is not a snapshot taken before each campaign: the May run records store no code commit, and some run inputs changed between and during campaigns. The release README (``Code and skill state by campaign'') lists those changes by results root. \texttt{benchmark/REPRODUCIBILITY.md} describes credentialed reconstruction: obtain PhysioNet access to \mimic{} v3.1 and \eicu{} v2.0, place the source files in the documented directories, run \texttt{uv sync}, \texttt{m4 init mimic-iv}, \texttt{m4 init-derived mimic-iv}, \texttt{m4 init eicu}, and \texttt{benchmark/setup.py -{}-all -{}-verify}, then run schema-equivalence verification, preflight, and the Docker leak canary before any fresh campaign.

\paragraph{Open evidence packages.}
Table~\ref{tab:evidence_packages} lists the two open packages. The packager hashes every source file against its declared manifest before transforming it and aborts on any missing, extra, or mismatched file. Each \texttt{output.csv} becomes \texttt{output\_summary.json} (columns, row count, source hash), and tool and command outputs in traces become placeholders with length and a keyed hash prefix. Agent messages, SQL, and code are kept. In that text, credentials, shifted datetimes, patient ages, and rows that start with an identifier are removed, and \mimic{} and \eicu{} identifiers are pseudonymized. Agent messages can still describe in words what the agent saw. A separate scanner with its own patterns checks each package for residual tool output, credentials, identifiers, and shifted datetimes, and both packages pass it. \texttt{metadata/public\_artifact\_manifest.json} maps each source hash to its public hash, and Croissant core and RAI metadata describe both packages. In the submitted review archive, 322 \texttt{result.json} files and \resultnum{2,513} trace events were malformed because packaging-time path redaction consumed JSON escape characters inside agent text. Evaluator fields were unaffected. The historical package uses documented derived copies of the \texttt{result.json} files and records original, derived, and public hashes. The malformed trace events are withheld whole, because their tool output cannot be separated from agent text. Original bytes are unchanged in the private archive.

\begin{table}[t]
\centering
\caption{Open evidence packages, hosted at \url{https://huggingface.co/datasets/hannesill/m4bench-evidence} and attached to the tagged release.}
\label{tab:evidence_packages}
\begin{tabular}{p{0.38\textwidth}p{0.28\textwidth}rr}
\toprule
Archive & Contents & Runs & Bytes \\
\midrule
\texttt{m4bench-evidence-}\newline\texttt{submission.tar.zst} & \raggedright Submitted campaigns (Codex primary, Codex follow-up, Claude sentinel, OSS) & 1,400 & 7,548,717 \\
\multicolumn{4}{l}{SHA-256: \texttt{f8b72940bd98cd9b2a852fa79c416310e2ed8ec7795b181941358d3b03ca708c}} \\
\midrule
\texttt{m4bench-evidence-}\newline\texttt{claude-replication.tar.zst} & \raggedright Full-coverage Claude replication & 560 & 6,398,244 \\
\multicolumn{4}{l}{SHA-256: \texttt{4aeb5eda83090f89e5a3bf5d0e8c4568c18309efd66b496c4655e11abd13471b}} \\
\bottomrule
\end{tabular}

\end{table}

\paragraph{Regeneration from packaged evidence.}
From a clean checkout of the tag, \texttt{verify\_public\_evidence.py} checks every package entry against its manifest and regenerates tables from the package alone. All four Claude replication tables, their summary file, and 18 of 19 historical result tables reproduce byte for byte. Historical key F1 uses the reference row counts in Table~\ref{tab:task_details}, and ordered-value nMAE uses the tracked per-run cache. Table~\ref{tab:codex_schema_skill_generalization} cannot be regenerated from the packages: its transformed-schema \noskill{} cells come from the April 2026 campaign root \texttt{codex-full-20260428}, which is not in either archive. Task-detail, artifact, and compute tables need the private archives or credentialed reference outputs, and re-evaluating agent outputs needs both.

\paragraph{Provenance limits.}
Runs record task, condition, model, trial, timestamps, isolation and canary checks, evaluator diagnostics, and database fingerprints, but not the source commit, working-tree state, or container image digest. For the Claude replication, the collaborator declares base commit \texttt{f2f0299d} plus an uncommitted diff that changes retry handling and task selection but not task definitions or the evaluator. The diff, launch commands, and batch scripts are released, and 17 canonical runs predate the retry change. Local runs are tied to retained model-pinning patches and benchmark source hashes. These are declarations and retained source history, not per-run records. A separate authentication-failure archive from the collaborator campaign was not retrieved, so its failed attempts are missing from the attempt inventory. It contains no canonical runs.

\section{Additional result details}
\label{sec:appendix_results_details}

\paragraph{Primary Codex campaign integrity.}
The final Codex evidence package combines \texttt{release-20260502-codex-v11} with \resultnum{170 clinician-review rerun replacements} from \texttt{review-rerun-20260504-codex}. After treating legacy SQL-free variant runs as the canonical \withskill{} rows for the four affected task variants, the analysis set contains \resultnum{760 runs} across \resultnum{152 task/condition/schema/model cells}. Those four variants are the standard and raw urine-output-rate and vasopressor-equivalent tasks. Their SQL-free skill text marked each stripped SQL fragment with a placeholder. On 5 May 2026 (commit \texttt{5f3ac09c}) the placeholders were replaced by written rules, and that later text is the released skill and the one the Codex follow-up and the Claude replication used. For these four tasks the primary Codex and Claude \withskill{} runs therefore did not see the same skill text. Publishability establishes eligibility for analysis, not clinical correctness: most failed diagnostics are valid failed or partially correct derivations, which is why the scalar reward is interpreted alongside key precision, extra-key counts, and strict pass/fail status.

\begin{table}[t]
\centering
\caption{Primary Codex campaign integrity checks after replacing matching May 2 runs with the May 4 clinician-review reruns. Publishability means the run is eligible for analysis. It does not mean the submitted derivation is clinically correct.}
\label{tab:codex_integrity}
\begin{tabular}{lr}
\toprule
Audit quantity & Count \\
\midrule
Runs & 760 \\
Expected primary/supplementary runs & 760 \\
Task/condition/schema/model cells & 152 \\
Publishable flag failures & 0 \\
Isolation flag failures & 0 \\
Filesystem canary failures & 0 \\
Contamination-lint failures & 0 \\
Nonzero agent exits & 3 \\
Missing rewards & 0 \\
Fully passing diagnostic runs & 87 \\
Runs with semantic diagnostic failures & 670 \\
Retained output.csv artifacts & 760 \\
Retained trace.jsonl artifacts & 760 \\
Retained instruction.md artifacts & 760 \\
Retained egress.jsonl artifacts & 760 \\
\bottomrule
\end{tabular}

\end{table}

\begin{figure}[t]
  \centering
  \includegraphics[width=0.92\textwidth]{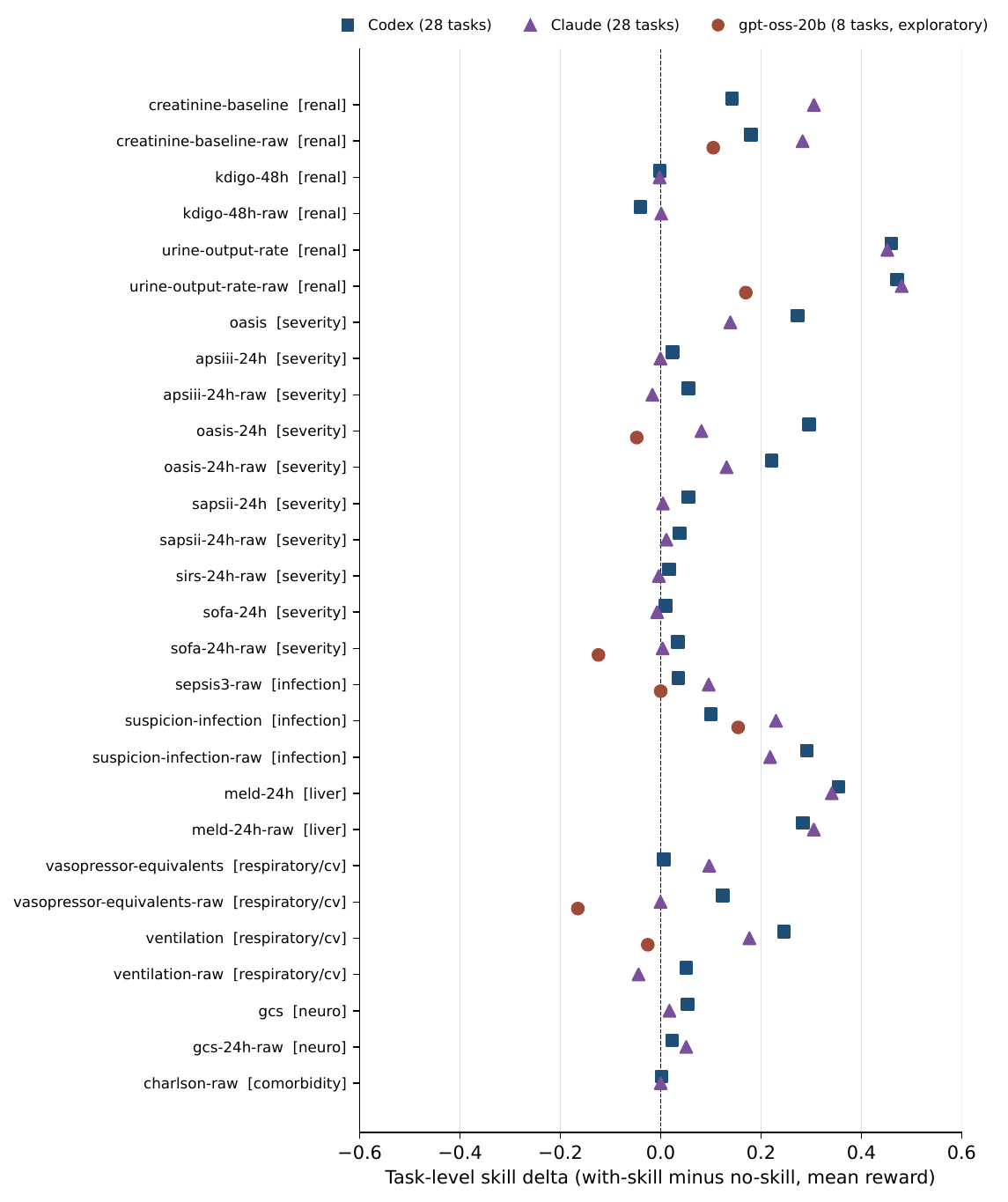}
  \caption{Per-task native-schema skill delta (with-skill minus no-skill mean reward) for the primary Codex campaign and the full-coverage Claude replication on all 28 tasks, and for the exploratory \texttt{gpt-oss-20b} arm on the eight tasks it covers.}
  \label{fig:skill_delta_forest}
\end{figure}

\paragraph{Codex heterogeneity.}
Figure~\ref{fig:skill_delta_forest} and Tables~\ref{tab:codex_skill_gains}--\ref{tab:codex_mode_skill_effect} detail where the primary effect is largest, where it is flat or negative, and how it splits by standard versus raw tasks.

\begin{table}[t]
\centering
\caption{Largest native-schema skill gains after averaging task-level deltas across the Codex models.}
\label{tab:codex_skill_gains}
\begin{tabular}{lr}
\toprule
Task & Mean delta \\
\midrule
mimic-urine-output-rate-raw & 0.471 \\
mimic-urine-output-rate & 0.460 \\
mimic-meld-24h & 0.354 \\
mimic-oasis-24h & 0.296 \\
mimic-suspicion-infection-raw & 0.291 \\
mimic-meld-24h-raw & 0.283 \\
eicu-oasis & 0.272 \\
mimic-ventilation & 0.246 \\
mimic-oasis-24h-raw & 0.221 \\
mimic-creatinine-baseline-raw & 0.180 \\
\bottomrule
\end{tabular}

\end{table}

\begin{table}[t]
\centering
\caption{Largest negative or flat native-schema skill effects after averaging task-level deltas across the Codex models.}
\label{tab:codex_skill_declines}
\begin{tabular}{lr}
\toprule
Task & Mean delta \\
\midrule
mimic-kdigo-48h-raw & -0.040 \\
mimic-kdigo-48h & -0.002 \\
mimic-charlson-raw & 0.001 \\
mimic-vasopressor-equivalents & 0.006 \\
mimic-sofa-24h & 0.010 \\
mimic-sirs-24h-raw & 0.016 \\
mimic-gcs-24h-raw & 0.023 \\
mimic-apsiii-24h & 0.023 \\
\bottomrule
\end{tabular}

\end{table}

\begin{table}[t]
\centering
\caption{Native-schema Codex skill effect split by task mode. Values are task-paired deltas. Final rows average the two Codex models within task before aggregating.}
\label{tab:codex_mode_skill_effect}
\begin{tabular}{llrrr}
\toprule
Model & Mode & Tasks & Mean delta & Positive tasks \\
\midrule
GPT 5.4 Mini & standard & 11 & 0.203 & 10/11 \\
GPT 5.4 Mini & raw & 17 & 0.136 & 16/17 \\
GPT 5.5 & standard & 11 & 0.105 & 9/11 \\
GPT 5.5 & raw & 17 & 0.112 & 15/17 \\
Task-balanced mean & standard & 11 & 0.154 & 10/11 \\
Task-balanced mean & raw & 17 & 0.124 & 16/17 \\
\bottomrule
\end{tabular}

\end{table}

\begin{table}[t]
\centering
\caption{Cluster sensitivity for the primary native-schema Codex skill effect. Task rows are the paper headline. Family rows resample or balance the 15 task families to check sensitivity to standard/raw sibling tasks.}
\label{tab:codex_family_sensitivity}
\begin{tabular}{lrrr}
\toprule
Analysis & Units & Mean delta & 95\% interval \\
\midrule
Task-cluster bootstrap & 28 tasks & 0.136 & [0.085, 0.191] \\
Family-cluster bootstrap & 15 families & 0.136 & [0.065, 0.207] \\
Family-balanced bootstrap & 15 families & 0.120 & [0.057, 0.190] \\
\bottomrule
\end{tabular}

\end{table}

\paragraph{Output-grain diagnostics.}
The diagnostic table reports aggregate key F1, key precision, and the largest extra-key cells, which are dominated by interval-style derivations.

\begin{table}[t]
\centering
\caption{Native-schema key diagnostics for Codex. The first two rows summarize native no-skill and with-skill cells. Subsequent rows show high-extra-key cells. N counts cells in the first two rows and runs in the others. Key F1 and key precision are means.}
\label{tab:codex_key_precision}
\resizebox{\textwidth}{!}{\begin{tabular}{lrrrr}
\toprule
Condition or worst cell & N & Key F1 & Key precision & Extra keys \\
\midrule
no-skill & 56 & 0.962 & 0.967 & 2384458 \\
with-skill & 56 & 0.965 & 0.976 & 2087534 \\
mimic-ventilation-raw / no-skill / GPT 5.4 Mini & 5 & 0.700 & 0.559 & 555030 \\
mimic-ventilation-raw / no-skill / GPT 5.5 & 5 & 0.717 & 0.573 & 525311 \\
mimic-ventilation-raw / with-skill / GPT 5.4 Mini & 5 & 0.720 & 0.583 & 505933 \\
mimic-ventilation-raw / with-skill / GPT 5.5 & 5 & 0.740 & 0.603 & 459225 \\
mimic-vasopressor-equivalents-raw / with-skill / GPT 5.4 Mini & 5 & 0.949 & 0.910 & 384994 \\
\bottomrule
\end{tabular}
}
\end{table}

\begin{table}[t]
\centering
\caption{Per-task secondary diagnostics for native-schema Codex. nMAE is computed only for ordered value columns after matching on declared keys. Negative nMAE deltas are improvements.}
\label{tab:codex_secondary_diagnostics_by_task}
\resizebox{\textwidth}{!}{\begin{tabular}{lrrrr}
\toprule
Task & Reward delta & Key F1 delta & nMAE delta & Ordered cols \\
\midrule
eicu-gcs & 0.054 & 0.000 & -0.029 & 4 \\
eicu-oasis & 0.272 & 0.000 & -0.189 & 11 \\
mimic-apsiii-24h & 0.023 & 0.000 & -0.013 & 17 \\
mimic-apsiii-24h-raw & 0.055 & 0.000 & -0.036 & 17 \\
mimic-charlson-raw & 0.001 & 0.000 & -0.001 & 19 \\
mimic-creatinine-baseline & 0.142 & -0.073 & 0.015 & 3 \\
mimic-creatinine-baseline-raw & 0.180 & 0.000 & 0.026 & 3 \\
mimic-gcs-24h-raw & 0.023 & 0.000 & -0.015 & 4 \\
mimic-kdigo-48h & -0.002 & 0.000 & -0.000 & 4 \\
mimic-kdigo-48h-raw & -0.040 & 0.000 & 0.031 & 4 \\
mimic-meld-24h & 0.354 & 0.000 & -0.219 & 1 \\
mimic-meld-24h-raw & 0.283 & 0.000 & -0.244 & 1 \\
mimic-oasis-24h & 0.296 & 0.000 & -0.116 & 11 \\
mimic-oasis-24h-raw & 0.221 & 0.000 & -0.102 & 11 \\
mimic-sapsii-24h & 0.055 & 0.000 & -0.034 & 16 \\
mimic-sapsii-24h-raw & 0.038 & 0.000 & -0.022 & 16 \\
mimic-sepsis3-raw & 0.035 & -0.089 & -0.102 & 8 \\
mimic-sirs-24h-raw & 0.016 & 0.000 & -0.011 & 5 \\
mimic-sofa-24h & 0.010 & 0.000 & -0.007 & 7 \\
mimic-sofa-24h-raw & 0.034 & 0.000 & -0.028 & 7 \\
mimic-suspicion-infection & 0.100 & -0.001 & -0.046 & 2 \\
mimic-suspicion-infection-raw & 0.291 & 0.140 & -0.092 & 2 \\
mimic-urine-output-rate & 0.460 & 0.000 & -0.072 & 6 \\
mimic-urine-output-rate-raw & 0.471 & 0.002 & -0.142 & 6 \\
mimic-vasopressor-equivalents & 0.006 & 0.000 & -0.000 & 1 \\
mimic-vasopressor-equivalents-raw & 0.123 & 0.005 & -0.000 & 1 \\
mimic-ventilation & 0.246 & 0.075 & NA & 0 \\
mimic-ventilation-raw & 0.051 & 0.021 & NA & 0 \\
\bottomrule
\end{tabular}
}
\end{table}

\begin{table}[t]
\centering
\caption{Matched-context controls on selected tasks in the final combined Codex analysis. ``Control--targeted'' compares the control condition to the targeted \withskill{} condition after task-level averaging.}
\label{tab:codex_matched_controls}
\resizebox{\textwidth}{!}{\begin{tabular}{lrrrrrr}
\toprule
Control & Tasks & No skill & Targeted & Control & Control--targeted & Control--no skill \\
\midrule
Raw reference SQL & 10 & 0.553 & 0.833 & 0.837 & 0.003 & 0.283 \\
Decoy skill & 10 & 0.553 & 0.833 & 0.554 & -0.280 & 0.000 \\
\bottomrule
\end{tabular}
}
\end{table}

\begin{table}[t]
\centering
\caption{Codex schema-skill generalization follow-up for \gptmini{} over 26 MIMIC tasks. Transformed rows match May 6 \withskill{} runs to the earlier transformed-schema \noskill{} probe.}
\label{tab:codex_schema_skill_generalization}
\resizebox{\textwidth}{!}{\begin{tabular}{lrrrrrr}
\toprule
Schema & Tasks & No skill & With skill & Skill delta & Positive tasks & Robustness ratio \\
\midrule
native & 26 & 0.637 & 0.794 & 0.157 [0.092, 0.229] & 24/26 & 1.000 \\
obfuscated & 26 & 0.561 & 0.696 & 0.135 [0.068, 0.208] & 19/26 & 0.862 \\
restructured & 26 & 0.557 & 0.668 & 0.110 [0.057, 0.170] & 23/26 & 0.704 \\
\bottomrule
\end{tabular}
}
\end{table}

\paragraph{Validity follow-up details.}
The May 6 Codex validity follow-up contains \resultnum{363 raw result directories}. The canonical analysis manifest selects the latest timestamped run for each task, condition, schema, model, and trial key, yielding \resultnum{340 canonical runs} and \resultnum{23 superseded retry directories}. The canonical matrix contains \resultnum{80 operational-spec runs} and \resultnum{260 transformed-schema with-skill runs}. All canonical runs pass publishability, isolation, filesystem-canary, and contamination-lint checks. The follow-up is used as validity evidence, not as a replacement for the primary native-schema estimand.

\begin{table}[t]
\centering
\caption{May 6 Codex validity follow-up integrity. Superseded retry directories are retained in the manifest for audit but excluded from analysis tables.}
\label{tab:codex_followup_integrity}
\begin{tabular}{lr}
\toprule
Audit quantity & Count \\
\midrule
Raw result directories & 363 \\
Canonical run keys & 340 \\
Superseded duplicate directories & 23 \\
Operational-spec canonical runs & 80 \\
Transformed-schema with-skill canonical runs & 260 \\
Publishable flag failures & 0 \\
Isolation flag failures & 0 \\
Filesystem canary failures & 0 \\
Contamination-lint failures & 0 \\
Nonzero agent exits & 3 \\
Fully passing diagnostic runs & 22 \\
operational-spec / native / GPT 5.4 Mini & 40 \\
operational-spec / native / GPT 5.5 & 40 \\
with-skill / obfuscated / GPT 5.4 Mini & 130 \\
with-skill / restructured / GPT 5.4 Mini & 130 \\
\bottomrule
\end{tabular}

\end{table}

\begin{table}[t]
\centering
\caption{Operational-spec follow-up by sentinel task after averaging the two Codex models within each task.}
\label{tab:codex_operational_spec_by_task}
\resizebox{\textwidth}{!}{\begin{tabular}{lrrrrr}
\toprule
Task & No skill & Operational spec & Targeted skill & Spec--no & Targeted--spec \\
\midrule
mimic-urine-output-rate-raw & 0.372 & 0.802 & 0.843 & 0.430 & 0.041 \\
mimic-ventilation & 0.536 & 0.528 & 0.782 & -0.008 & 0.254 \\
mimic-creatinine-baseline-raw & 0.620 & 0.794 & 0.800 & 0.174 & 0.006 \\
mimic-suspicion-infection & 0.720 & 0.932 & 0.820 & 0.212 & -0.112 \\
mimic-vasopressor-equivalents-raw & 0.733 & 0.842 & 0.856 & 0.109 & 0.014 \\
mimic-oasis-24h & 0.699 & 0.987 & 0.995 & 0.288 & 0.008 \\
mimic-sepsis3-raw & 0.356 & 0.400 & 0.391 & 0.044 & -0.009 \\
mimic-sofa-24h-raw & 0.812 & 0.854 & 0.846 & 0.041 & -0.007 \\
\bottomrule
\end{tabular}
}
\end{table}

\paragraph{Historical Claude sentinel and exploratory OSS arms.}
The submitted Claude sentinel campaign, superseded for provider-scope claims by the full-coverage replication in Appendix~\ref{sec:appendix_claude28} and retained as historical evidence, combines \texttt{release-20260502-claude-provider} with \resultnum{30 May 4 rerun replacements}, yielding \resultnum{190 runs and 14 strict diagnostic passes}. Claude schema-perturbation evidence remains inconclusive unless a future rerun includes matched native, obfuscated, and restructured cells for the same tasks. The local \texttt{gpt-oss-20b} arm is useful for reproducibility context but not paper-facing evidence under our release criteria: all \resultnum{110 runs} are marked non-publishable because the local Ollama host exception was enabled, no run passes every diagnostic, and no \texttt{egress.jsonl} artifacts are retained.

\begin{table}[t]
\centering
\caption{Historical (submitted) Claude sentinel native-schema skill effect across eight tasks.}
\label{tab:claude_native_skill_effect}
\begin{tabular}{lrrrr}
\toprule
Model & No skill & With skill & Task-paired delta & Positive tasks \\
\midrule
Claude Opus & 0.738 & 0.843 & 0.106 & 6/8 \\
Claude Sonnet & 0.592 & 0.817 & 0.226 & 8/8 \\
Task-balanced mean & -- & -- & 0.166 [0.085, 0.259] & 8/8 \\
\bottomrule
\end{tabular}

\end{table}

\begin{table}[t]
\centering
\caption{Historical (submitted) Claude sentinel per-task native-schema skill deltas.}
\label{tab:claude_task_deltas}
\begin{tabular}{lrrr}
\toprule
Task & Opus delta & Sonnet delta & Mean delta \\
\midrule
urine-output rate raw & 0.391 & 0.484 & 0.438 \\
baseline creatinine raw & 0.160 & 0.298 & 0.229 \\
ventilation & 0.159 & 0.263 & 0.211 \\
OASIS-24h & 0.103 & 0.214 & 0.158 \\
suspected infection & 0.043 & 0.242 & 0.143 \\
vasopressor equivalents raw & -0.021 & 0.219 & 0.099 \\
Sepsis-3 raw & 0.024 & 0.070 & 0.047 \\
SOFA-24h raw & -0.013 & 0.016 & 0.001 \\
\bottomrule
\end{tabular}

\end{table}

\begin{table}[t]
\centering
\caption{Exploratory local OSS native-schema skill effect. This arm is excluded from primary and supplementary paper-facing evidence because local Ollama runs do not meet the release publishability criteria.}
\label{tab:oss_native_skill_effect}
\begin{tabular}{lrrrr}
\toprule
Model & No skill & With skill & Task-paired delta & Positive tasks \\
\midrule
gpt-oss-20b local & 0.085 & 0.094 & 0.008 [-0.072, 0.088] & 3/8 \\
\bottomrule
\end{tabular}

\end{table}

\section{Full-coverage Claude replication details}
\label{sec:appendix_claude28}

\paragraph{Selection and sources.}
The canonical 560 runs combine 11 tasks completed in a local root (\texttt{rebuttal-claude-opus-sonnet-20260726}, 220 runs), seven disjoint tasks from a second local root (\texttt{rebuttal-claude-opus-sonnet-additional-20260728}, 140 runs), and ten tasks completed by a collaborator (\texttt{claude-28-20260726}, 200 runs, newest attempt per cell). Every canonical run is publishable, isolated, and passes the filesystem canary and contamination lint, with all five artifacts present and hashed. Excluded attempts are inventoried: \resultnum{119} runs quarantined after authentication was revoked, \resultnum{463} runtime failures or incomplete duplicates, \resultnum{57} overlapping local runs outside the declared selection, \resultnum{16} superseded collaborator attempts, and one isolation canary. Earlier interim estimates reported during discussion used a fixed-cutoff 15-task subset, and the complete manifest supersedes them. Per-run reward is taken from the retained evaluator records, not from rounded cell summaries. As a descriptive cross-cohort check, \resultnum{11} collaborator cells were also completed locally with five runs each (within the \resultnum{57} overlapping runs): the mean absolute difference between their cell means and the canonical ones is \resultnum{0.021}, the largest \resultnum{0.052}, and substituting the local runs for the two fully duplicated tasks moves the 28-task delta from \resultnum{+0.1196} to \resultnum{+0.1205} (computed from the released canonical manifest and attempt inventory). Both duplicated tasks have near-zero skill deltas, so this is a weak check of numerical stability across the two machines and says nothing about harness equivalence.

\paragraph{Harness profile.}
Table~\ref{tab:claude28_harness_profile} counts the recorded harness indicators by condition. All collaborator runs used Claude Code 2.1.119. The local runs used 2.1.220 except 67 \noskill{} runs on 2.1.119. Eight task comparisons mix versions, and raw SAPS~II has recorded model-slot overrides in \noskill{} but not in \withskill{}. These nine tasks are excluded from the matched subset in Table~\ref{tab:claude28_sensitivity}. Requested Opus or Sonnet was the primary assistant model in every run (the collaborator runs requested the \texttt{opus} and \texttt{sonnet} aliases, which resolved to the same model identifiers). Ancillary Haiku usage is a property of Claude Code native routing, so these runs are Claude Code with requested Opus/Sonnet, not pure single-model executions. Resolved reasoning effort was medium for every run.

\begin{table}[t]
\centering
\caption{Full-coverage Claude replication: model-specific and task-balanced native-schema \withskill{} minus \noskill{} reward differences over 28 tasks (95\% task-bootstrap intervals).}
\label{tab:claude28_native_skill_effect}
\begin{tabular}{lrrrr}
\toprule
Model & No skill & With skill & Task-paired delta & Positive tasks \\
\midrule
Claude Opus & 0.800 & 0.883 & 0.083 [0.038, 0.135] & 25/28 \\
Claude Sonnet & 0.694 & 0.851 & 0.156 [0.087, 0.231] & 20/28 \\
Task-balanced mean & 0.747 & 0.867 & 0.120 [0.067, 0.177] & 21/28 \\
\bottomrule
\end{tabular}

\end{table}

\begin{table}[t]
\centering
\caption{Recorded harness indicators of the full-coverage Claude replication by condition: observed Claude Code version, recorded model-slot overrides, runs whose usage reports Haiku, runs with Haiku assistant turns, and summed agent wall-clock hours. Hardware was not recorded per run.}
\label{tab:claude28_harness_profile}
\resizebox{\textwidth}{!}{\begin{tabular}{lrrrrrrr}
\toprule
Condition & Runs & CLI 2.1.119 & CLI 2.1.220 & Overrides recorded & Haiku usage & Haiku turns & Agent hours \\
\midrule
no-skill & 280 & 167 & 113 & 120 & 160 & 4 & 28.6 \\
with-skill & 280 & 100 & 180 & 110 & 170 & 1 & 22.6 \\
\bottomrule
\end{tabular}
}
\end{table}

\begin{table}[t]
\centering
\caption{Full-coverage Claude replication per-task skill deltas. ``Harness'' marks tasks whose \noskill{} and \withskill{} cells match in recorded CLI version and routing profile, or the indicator that differs.}
\label{tab:claude28_task_deltas}
\resizebox{0.8\textwidth}{!}{\begin{tabular}{lrrrl}
\toprule
Task & Opus delta & Sonnet delta & Mean delta & Harness \\
\midrule
mimic-urine-output-rate-raw & 0.476 & 0.485 & 0.480 & matched \\
mimic-urine-output-rate & 0.415 & 0.488 & 0.452 & matched \\
mimic-meld-24h & 0.067 & 0.615 & 0.341 & CLI \\
mimic-creatinine-baseline & 0.232 & 0.379 & 0.305 & matched \\
mimic-meld-24h-raw & 0.165 & 0.445 & 0.305 & CLI \\
mimic-creatinine-baseline-raw & 0.160 & 0.406 & 0.283 & matched \\
mimic-suspicion-infection & 0.169 & 0.291 & 0.230 & CLI \\
mimic-suspicion-infection-raw & 0.225 & 0.211 & 0.218 & matched \\
mimic-ventilation & -0.056 & 0.409 & 0.177 & matched \\
eicu-oasis & 0.129 & 0.149 & 0.139 & matched \\
mimic-oasis-24h-raw & 0.096 & 0.167 & 0.131 & CLI \\
mimic-vasopressor-equivalents & 0.083 & 0.110 & 0.097 & matched \\
mimic-sepsis3-raw & 0.093 & 0.098 & 0.096 & matched \\
mimic-oasis-24h & 0.068 & 0.094 & 0.081 & CLI \\
mimic-gcs-24h-raw & 0.071 & 0.030 & 0.051 & matched \\
eicu-gcs & 0.003 & 0.032 & 0.017 & matched \\
mimic-sapsii-24h-raw & 0.000 & 0.023 & 0.011 & routing \\
mimic-sapsii-24h & 0.011 & -0.002 & 0.004 & CLI \\
mimic-sofa-24h-raw & 0.007 & 0.000 & 0.004 & matched \\
mimic-kdigo-48h-raw & 0.022 & -0.020 & 0.001 & matched \\
mimic-charlson-raw & 0.004 & -0.004 & 0.000 & matched \\
mimic-vasopressor-equivalents-raw & -0.089 & 0.088 & -0.000 & matched \\
mimic-apsiii-24h & 0.030 & -0.031 & -0.000 & matched \\
mimic-kdigo-48h & 0.001 & -0.006 & -0.002 & matched \\
mimic-sirs-24h-raw & 0.010 & -0.018 & -0.004 & CLI \\
mimic-sofa-24h & 0.015 & -0.029 & -0.007 & CLI \\
mimic-apsiii-24h-raw & 0.017 & -0.049 & -0.016 & matched \\
mimic-ventilation-raw & -0.105 & 0.017 & -0.044 & matched \\
\bottomrule
\end{tabular}
}
\end{table}

\clearpage
\section*{NeurIPS Paper Checklist}

\begin{enumerate}

\item {\bf Claims}
    \item[] Question: Do the main claims made in the abstract and introduction accurately reflect the paper's contributions and scope?
    \item[] Answer: \answerYes{}.
    \item[] Justification: The abstract and introduction frame \mbench{} as an evaluation benchmark for reference EHR derivation agreement, not as validation of autonomous clinical correctness. The main claims are bounded to curated procedural context under the declared benchmark conditions, and the cross-provider Claude replication is presented as descriptive because its harness was not uniform.

\item {\bf Limitations}
    \item[] Question: Does the paper discuss the limitations of the work performed by the authors?
    \item[] Answer: \answerYes{}.
    \item[] Justification: Section~6 discusses what the benchmark measures, interpretation limits and non-claims, statistical and external validity, and safety boundaries; Sections~1 and~3 state the scope of clinician review and of the \noskill{} baseline.

\item {\bf Theory assumptions and proofs}
    \item[] Question: For each theoretical result, does the paper provide the full set of assumptions and a complete (and correct) proof?
    \item[] Answer: \answerNA{}.
    \item[] Justification: The paper introduces an evaluation benchmark and empirical protocol; it does not present theoretical results.

    \item {\bf Experimental result reproducibility}
    \item[] Question: Does the paper fully disclose all the information needed to reproduce the main experimental results of the paper to the extent that it affects the main claims and/or conclusions of the paper (regardless of whether the code and data are provided or not)?
    \item[] Answer: \answerYes{}.
    \item[] Justification: Sections~3--4 and the appendix specify tasks, conditions, metrics, execution, isolation, result roots, artifact retention, and repeated-trial aggregation. Exact replay of individual agent trajectories may differ because coding agents and hosted model APIs are nondeterministic, so the paper reports task-balanced effects with bootstrap intervals and sign-flip tests, and releases per-run metadata, sanitized per-run evidence packages with hash lineage, canonical run manifests, and scripts that regenerate the result tables from those packages, except the schema-generalization table, whose \noskill{} cells come from runs outside the packages (Appendix~\ref{sec:appendix_release}).

\item {\bf Open access to data and code}
    \item[] Question: Does the paper provide open access to the data and code, with sufficient instructions to faithfully reproduce the main experimental results, as described in supplemental material?
    \item[] Answer: \answerYes{}.
    \item[] Justification: The tagged source release contains evaluation code, task specifications, skills, the isolation harness, credentialed reconstruction instructions, run manifests, and table scripts; open evidence packages with checksums and Croissant metadata contain per-run results and sanitized traces (Appendix~\ref{sec:appendix_release}). MIMIC-IV and eICU, derived task databases, reference outputs, and row-level agent outputs remain subject to PhysioNet credentialing and data-use terms, so data-layer reproduction requires independent credentialed access.

\item {\bf Experimental setting/details}
    \item[] Question: Does the paper specify all the training and test details (e.g., data splits, hyperparameters, how they were chosen, type of optimizer) necessary to understand the results?
    \item[] Answer: \answerYes{}.
    \item[] Justification: The study does not train models. Sections~3--4 describe benchmark construction, model/provider arms, conditions, trials, scoring, isolation, and recorded reporting fields.

\item {\bf Experiment statistical significance}
    \item[] Question: Does the paper report error bars suitably and correctly defined or other appropriate information about the statistical significance of the experiments?
    \item[] Answer: \answerYes{}.
    \item[] Justification: Section~4 defines the uncertainty procedure using task-cluster bootstrap intervals, family-cluster sensitivity analysis, and a Monte Carlo sign-flip test against zero. Section~5 and the appendix report bootstrap intervals for the relevant task-balanced effect estimates, and Section~4 states that bracketed intervals are 95\% non-parametric percentile bootstrap intervals over task-level deltas, not standard deviations or standard errors over repeated runs.

\item {\bf Experiments compute resources}
    \item[] Question: For each experiment, does the paper provide sufficient information on the computer resources (type of compute workers, memory, time of execution) needed to reproduce the experiments?
    \item[] Answer: \answerYes{}.
    \item[] Justification: Appendix Table~\ref{tab:compute_resources} reports hardware, retained-run wall-clock time, token usage, and retained artifact storage for the Codex primary and Claude sentinel campaigns. For the full-coverage Claude replication we report summed agent wall-clock time by condition (Appendix Table~\ref{tab:claude28_harness_profile}); hardware was not recorded per run. Compute for the Codex follow-up, the exploratory OSS arm, pilot runs, and excluded attempts is not reported.
    
\item {\bf Code of ethics}
    \item[] Question: Does the research conducted in the paper conform, in every respect, with the NeurIPS Code of Ethics \url{https://neurips.cc/public/EthicsGuidelines}?
    \item[] Answer: \answerYes{}.
    \item[] Justification: The benchmark uses credentialed de-identified clinical research databases under their data-use agreements, follows source-data access restrictions for derived databases and potentially data-bearing artifacts, and limits claims to supervised research derivations rather than clinical deployment.

\item {\bf Broader impacts}
    \item[] Question: Does the paper discuss both potential positive societal impacts and negative societal impacts of the work performed?
    \item[] Answer: \answerYes{}.
    \item[] Justification: Section~1 states the intended benefit, making clinical derivation knowledge versioned and auditable for supervised research workflows, and Section~6 discusses risks from incorrect derivations propagating into cohorts, risk adjustment, or downstream analyses.
    
\item {\bf Safeguards}
    \item[] Question: Does the paper describe safeguards that have been put in place for responsible release of data or models that have a high risk for misuse (e.g., pre-trained language models, image generators, or scraped datasets)?
    \item[] Answer: \answerYes{}.
    \item[] Justification: The release respects PhysioNet credentialing for MIMIC-IV and eICU data, releases only sanitized per-run evidence that withholds row-level outputs and tool outputs and is scanned for residual identifiers and shifted dates, keeps the original data-bearing archives private, and does not claim clinical deployment readiness.

\item {\bf Licenses for existing assets}
    \item[] Question: Are the creators or original owners of assets (e.g., code, data, models), used in the paper, properly credited and are the license and terms of use explicitly mentioned and properly respected?
    \item[] Answer: \answerYes{}.
    \item[] Justification: The paper cites MIMIC-IV, eICU, and MIT-LCP MIMIC-Code and names the evaluated agent products and model identifiers in Section~4. Section~3 and the appendix asset-license table report dataset versions, source-asset roles, PhysioNet credentialing/data-use terms and URLs, the M4Bench MIT release license, and access restrictions for derived databases and data-bearing artifacts.

\item {\bf New assets}
    \item[] Question: Are new assets introduced in the paper well documented and is the documentation provided alongside the assets?
    \item[] Answer: \answerYes{}.
    \item[] Justification: \mbench{} introduces task specifications, skills, evaluator code, and result artifacts. Sections~3--4 and the appendix document task inventory, clinician-review provenance, artifact retention, rerun requirements, and the evidence packages with checksums, licenses, and Croissant metadata.

\item {\bf Crowdsourcing and research with human subjects}
    \item[] Question: For crowdsourcing experiments and research with human subjects, does the paper include the full text of instructions given to participants and screenshots, if applicable, as well as details about compensation (if any)? 
    \item[] Answer: \answerNA{}.
    \item[] Justification: The paper does not report crowdsourcing experiments or prospective human-subjects experiments. Clinician task review is provenance and face-validity review of benchmark artifacts, not an experimental subject intervention.

\item {\bf Institutional review board (IRB) approvals or equivalent for research with human subjects}
    \item[] Question: Does the paper describe potential risks incurred by study participants, whether such risks were disclosed to the subjects, and whether Institutional Review Board (IRB) approvals (or an equivalent approval/review based on the requirements of your country or institution) were obtained?
    \item[] Answer: \answerNA{}.
    \item[] Justification: The work uses de-identified credentialed clinical research databases and benchmark artifact review; it does not recruit study participants or run interventions.

\item {\bf Declaration of LLM usage}
    \item[] Question: Does the paper describe the usage of LLMs if it is an important, original, or non-standard component of the core methods in this research? Note that if the LLM is used only for writing, editing, or formatting purposes and does \emph{not} impact the core methodology, scientific rigor, or originality of the research, declaration is not required.
    \item[] Answer: \answerYes{}.
    \item[] Justification: LLM coding agents are the evaluated systems. Sections~4--5 describe the model/provider arms, conditions, and how LLM agent outputs are scored.

\end{enumerate}

\end{document}